%% file: swd.tex
\documentclass[10pt,twocolumn,letterpaper]{article}

\usepackage{cvpr}
\usepackage{times}
\usepackage{epsfig}
\usepackage{graphicx}
\usepackage{amsmath}
\usepackage{amssymb}

\usepackage{booktabs} 
\usepackage{relsize} 
\usepackage{bbm} 
\newcommand*\rot{\rotatebox{90}}

\newcommand{\T}{{\mathcal{T}}}
   
\newcommand{\Tmus}{{\T_{\#}}{\mu}}
\newcommand{\x}{{\bf z}}
\newcommand{\dist}[1]{c(#1)}
\newcommand{\G}{{\boldsymbol{\gamma}}}
\newcommand{\proba}[1]{\mathcal{P}(#1)}
\newcommand{\Gzero}{{\boldsymbol{\gamma}}^*}

\usepackage{algorithm}%
\usepackage{algorithmic}
\usepackage[font=small]{caption}
\usepackage[pagebackref=true,breaklinks=true,letterpaper=true,colorlinks,bookmarks=false]{hyperref}

\cvprfinalcopy 

\def\cvprPaperID{4238} 

\ifcvprfinal\fi
\begin{document}

\title{Sliced Wasserstein Discrepancy for Unsupervised Domain Adaptation}

\author{Chen-Yu Lee \hspace{6mm} Tanmay Batra \hspace{6mm} Mohammad Haris Baig \hspace{6mm} Daniel Ulbricht\\
  Apple Inc\\
  }
  
\maketitle

\begin{abstract}
In this work, we connect two distinct concepts for unsupervised domain adaptation: feature distribution alignment between domains by utilizing the task-specific decision boundary~\cite{saito2017maximum} and the Wasserstein metric~\cite{villani2009optimal}. Our proposed sliced Wasserstein discrepancy (SWD) is designed to capture the natural notion of dissimilarity between the outputs of task-specific classifiers. It provides a geometrically meaningful guidance to detect target samples that are far from the support of the source and enables efficient distribution alignment in an end-to-end trainable fashion. In the experiments, we validate the effectiveness and genericness of our method on digit and sign recognition, image classification, semantic segmentation, and object detection.

\end{abstract}


\section{Introduction}

Deep convolutional neural networks~\cite{krizhevsky2012imagenet} is a milestone technique in the development of modern machine perception systems solving various tasks such as classification, semantic segmentation, object detection, etc.
However, in spite of the exceptional learning capacity and the improved generalizability, deep learning models still suffer from the challenge of domain shift -- a shift in the relationship between data collected in two different domains~\cite{ben2007analysis, ben2010theory} (\eg synthetic and real). 
Models trained on data collected in one domain can perform poorly on other domains.
Domain shift can exist in multiple forms: covariate shift (distribution shift in attributes), prior probability shift (shift in labels), and concept shift (shift in the relationship between attributes and labels)~\cite{shimodaira2000improving, storkey2009training, moreno2012unifying}.

In this paper, we focus on the covariate shift problem for the case where we have access to labeled data from one domain (source) and unlabeled data from another domain (target). This setup is commonly called unsupervised domain adaptation.
 Most of the work done in this field has focused on establishing a direct alignment between the feature distribution of source and target domains.
 Such alignment involves minimizing some distance measure of the feature distribution learned by the models~\cite{saenko2010adapting, ganin2014unsupervised, long2015learning}.
 More sophisticated methods use adversarial training~\cite{goodfellow2014generative} to further improve the quality of alignment between distributions by adapting representations at feature-level~\cite{hoffman2016fcns, ganin2016domain}, pixel-level~\cite{liu2016coupled, tzeng2017adversarial, bousmalis2017unsupervised}, or output-level~\cite{tsai2018learning} across domains.
 
A recent advance that moves beyond the direction of plain distribution matching was presented by Saito \etal in~\cite{saito2017maximum}.
They propose a within-network adversarial learning-based method containing a feature generator and two (task-specific) classifiers, which uses the task-specific decision boundary for aligning source and target samples.
Their method defines a new standard in developing generic domain adaptation frameworks. However, the system does have some limitations. For instance, their discrepancy loss ($L_1$ in this case) is only helpful when the two output probability measures from the classifiers overlap.

Inspired by the framework in~\cite{saito2017maximum}, we focus our efforts on improving the discrepancy measure which plays a central role in such within-network adversarial learning-based approach.
Our method aims to minimize the cost of moving the marginal distributions between the task-specific classifiers by utilizing the Wasserstein metric~\cite{monge1781memoire, kantorovitch1958translocation, arjovsky2017wasserstein}, which provides a more meaningful notion of dissimilarity for probability distributions.
We make several key contributions in this work:
(1) a novel and principled method for aligning feature distributions between domains via optimal transport theory (\ie,Wasserstein distance) and the task-specific decision boundary.
(2) enable efficient end-to-end training using sliced Wasserstein discrepancy (a variational formulation of Wasserstein metric).
(3) effectively harness the geometry of the underlying manifold created by optimizing the sliced Wasserstein discrepancy in an adversarial manner.
(4) the method advances the state-of-the-art across several tasks and can be readily applied to any domain adaptation problem such as image classification, semantic segmentation, and object detection.

\section{Related Work}
A rich body of approaches to unsupervised domain adaptation aim to reduce the gap between the source and target domains by learning domain-invariant feature representations, through various statistical moment matching techniques. Some methods utilize maximum mean discrepancy (MMD)~\cite{long2015learning, long2016unsupervised} to match the hidden representations of certain layers in a deep neural network. Other approaches uses Central moment discrepancy (CMD) method~\cite{zellinger2017central} to explicitly match each order and each hidden coordinate of higher order moments. Adaptive batch normalization (AdaBN)~\cite{li2018adaptive} has also been proposed to modulate the statistics in all batch normalization layers across the network between domains.

Another family of strategies tackles the domain adaptation problem by leveraging the adversarial learning behavior of GANs~\cite{goodfellow2014generative}. Such technique was first used at \textit{feature-level} where a domain discriminator is trained to correctly classify the domain of each input feature and the feature generator is trained to deceive the domain discriminator so that the resulting feature distribution is made domain invariant~\cite{tzeng2014deep, hoffman2016fcns, ganin2016domain}. Later the technique was applied at \textit{pixel-level} to perform distribution alignment in raw input space, translating source domain to the ``style'' of target domain and obtaining models trained on transformed source data~\cite{liu2016coupled, tzeng2017adversarial, bousmalis2017unsupervised, shrivastava2017learning, hoffman2017cycada, sankaranarayanan2017generate, murez2018image}. Recently the technique was used at \textit{output-level} by assuming that the output space contains similar spatial structure for some specific tasks such as semantic segmentation. The method in~\cite{tsai2018learning} thereby aligns pixel-level ground truth through adversarial learning in the output space. Other hybrid approaches have also been proposed in~\cite{sankaranarayanan2018learning, huang2018domain}.

In contrast, Saito \etal in~\cite{saito2017maximum} proposed to align distributions by explicitly utilizing task-specific classifiers as a discriminator. The framework maximizes the discrepancy between two classifiers' output to detect target samples that are outside the support of the source and then minimizes the discrepancy to generate feature representations that are inside the support of the source with respect to the decision boundary. Instead of aligning manifold in feature, input, or output space by heuristic assumptions, this approach focuses on directly reshaping the target data regions that indeed need to be reshaped.

Wasserstein metric, the natural geometry for probability measures induced by the optimal transport theory, has been investigated in several fields such as image retrieval~\cite{rubner1998metric}, color-based style transfer~\cite{pitie2005n}, and image warping~\cite{haker2004optimal}. The Wasserstein distance has also recently raised interest in stabilizing generative modeling~\cite{arjovsky2017wasserstein, deshpande2018generative, wu2018wasserstein}, learning introspective neural networks~\cite{lee2018wasserstein}, and obtaining Gaussian mixture models~\cite{kolouri2017sliced} thanks to its geometrically meaningful distance measure even when the supports of the distributions do not overlap.

As for domain adaptation, Courty \etal in~\cite{courty2015optimal} first learn a transportation plan matching source and target samples with class regularity. JDOT method~\cite{courty2017joint} learns to map input space from source to target by jointly considering the class regularity and feature distribution. DeepJDOT method~\cite{damodaran2018deepjdot} further improves upon JDOT by jointly matching feature and label space distributions with more discriminative feature representations in a deep neural network layer. However, the fact that these approaches explicitly enforce an one-to-one mapping between source samples and target samples in label space could largely restrict the practical usages when balanced source-target pairs are unavailable. It is also unclear how to extend these approaches to more generic tasks when one data sample has structured output space such as pixel-wise semantic segmentation.

In this work, we propose a principled framework to marry the two powerful concepts: distribution alignment by task-specific decision boundary~\cite{saito2017maximum} and the Wasserstein distance~\cite{villani2009optimal}. The Wasserstein metric serves as a reliable discrepancy measure between the task-specific classifiers, which directly measures the support of target samples from source samples instead of producing explicit one-to-one mapping in label space. A variational version of the Wasserstein discrepancy further provides straightforward and geomatrically meaningful gradients to jointly train the feature generator and classifiers in the framework efficiently.

\begin{figure*}
\begin{center}
\includegraphics[width=0.98\linewidth]{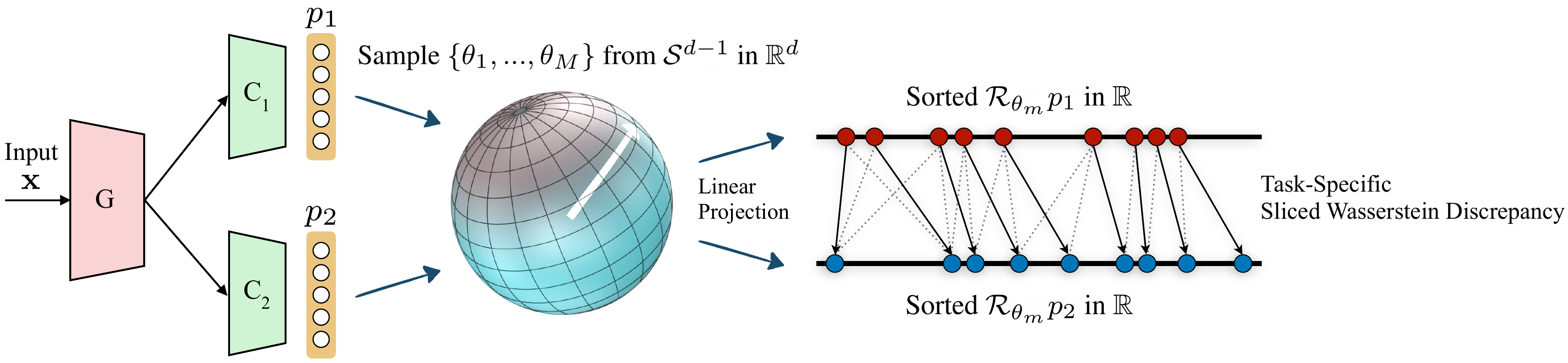}
\end{center}
\vspace{-5mm}
   \caption{An illustration of the proposed sliced Wasserstein discrepancy (SWD) computation. The SWD is designed to capture the dissimilarity of probability measures $p_1$ and $p_2$ in $\mathbb{R}^d$ between the task-specific classifiers $C_1$ and $C_2$, which take input from feature generator $G$. The SWD enables end-to-end training directly through a variational formulation of Wasserstein metric using radial projections on the uniform measures on the unit sphere $\mathcal{S}^{d-1}$, providing a geometrically meaningful guidance to detect target samples that are far from the support of the source. Please refer to Section~\ref{sec:wd} for details.}
\label{fig:pipeline}
\end{figure*}

\section{Method}
We first introduce unsupervised domain adaptation setting in Section~\ref{sec:setting}. Second, we briefly review the concept of optimal transport in Section~\ref{sec:ot}. Finally, we detail how to train the proposed method with the sliced Wasserstein discrepancy in Section~\ref{sec:wd}. 

\subsection{Framework Setup}
\label{sec:setting}

Given input data $\mathbf{x_{s}}$ and the corresponding ground truth $\mathbf{y_{s}}$ drawn from the source set \{$X_{s}, Y_{s}$\}, and input data $\mathbf{x_{t}}$ drawn from the target set $X_{t}$, the goal of unsupervised domain adaptation is to establish knowledge transfer from the labeled source set to the unlabeled target set as mentioned in~\cite{pan2010survey}.
 When the two data distributions $X_{s}$ and $X_{t}$ are close enough, one can simply focus on minimizing an empirical risk of the joint probability distribution $\mathcal{P} (X_{s}, Y_{s})$. However, when that the two distributions are substantially different, optimizing a model solely over the source information results in poor generalizability.

Following the Maximum Classifier Discrepancy (MCD) framework~\cite{saito2017maximum}, we train a feature generator network $G$ and
the classifier networks $C_1$ and $C_2$, which take feature responses generated from $G$ and produce the corresponding logits $p_1(\mathbf{y}|\mathbf{x})$, $p_2(\mathbf{y}|\mathbf{x})$ respectively (as shown in Figure~\ref{fig:pipeline}).
The optimization procedure consists of three steps:

\noindent
(1) train both generator $G$ and classifiers $(C_1, C_2)$ on the source domain \{$X_{s}, Y_{s}$\} to classify or regress the source samples correctly,
 \begin{equation}
 \small
   \min_{G, C_1, C_2} \mathcal{L}_s (X_{s},Y_{s}),
 \end{equation}
where $ \mathcal{L}_s$ can be any loss functions of interest such as cross entropy loss or mean squared error loss.

\noindent
(2) freeze the parameters of the generator $G$ and update the classifiers $(C_1, C_2)$ to maximize the discrepancy between the outputs of the two classifiers on the target set $X_{t}$, identifying the target samples that are outside the support of task-specific decision boundaries, 
 \begin{equation}
  \small
  \min_{C_1, C_2} \mathcal{L}_s (X_{s},Y_{s}) - \mathcal{L}_{\text{DIS}}(X_{t})
 \end{equation}
 where $ \mathcal{L}_{\text{DIS}}(X_{t})$ is the discrepancy loss ($L_1$ in~\cite{saito2017maximum}). $\mathcal{L}_s (X_{s},Y_{s})$ is also added to this step to retain information from the source domain, and

\noindent
(3) freeze the parameters of the two classifiers and update the generator $G$ to minimize the discrepancy between the outputs of the two classifiers on the target set $X_{t}$,
 \begin{equation}
   \small
  \min_{G} \mathcal{L}_{\text{DIS}}(X_{t})
 \end{equation}
 This step brings the target feature manifold closer to the source.

\subsection{Optimal Transport and Wasserstein Distance}
\label{sec:ot}
The effectiveness of domain adaptation in the aforementioned MCD framework depends entirely on the reliability of the discrepancy loss. Learning without the discrepancy loss, essentially dropping step 2 and step 3 in the training procedure, is simply supervised learning on the source domain. 

The Wasserstein distance has recently received great attention in designing loss functions for its superiority over other probability measures~\cite{wu2018wasserstein, mi2018variational}.
In comparison to other popular probability measures such as total variation distance, Kullback-Leibler divergence, and Jensen-Shannon divergence that compare point-wise histogram embeddings alone, Wasserstein distance takes into account the properties of the underlying geometry of probability space and it is even able to compare distribution measures that do not share support~\cite{arjovsky2017wasserstein}.
Motivated by the advantages of the Wasserstein distance, we now describe how we leverage this metric for measuring the discrepancy in our method.

Let $\Omega$ be a probability space and $\mu$, $\nu$ be two probability measures in $\mathcal{P}(\Omega)$, the Monge problem~\cite{monge1781memoire} seeks a transport map $\T:\Omega\rightarrow\Omega$ that minimizes the cost
\begin{equation}
\label{monge}
\small
\inf_{\Tmus=\nu}  \int_{\Omega} \dist{\x,\T(\x)}d\mu(\x),
\end{equation}
where  $\Tmus=\nu$ denotes a one-to-one push-forward from $\mu$ toward $\nu$ $\forall \text{ Borel subset A} \subset \Omega$ and $c: \Omega \times \Omega \rightarrow \mathbb{R}^+$ is a geodesic metric that can be either linear or quadratic. However, the  solution $\T^*$ may not always exist due to the assumption of no splitting of the probability measures, for example when pushing a Dirac measure toward a non-Dirac measure.

Kantorovitch~\cite{kantorovitch1958translocation} proposed a relaxed version of Eq~\ref{monge}, which seeks a transportation plan of a joint probability distribution $\G \in \proba{\Omega \times \Omega}$ such that
\begin{equation}
\small
\inf_{\G \in \Pi(\mu,\nu)}  \int_{\Omega\times\Omega} c(\x_1,\x_2)d\G(\x_1,\x_2),
\end{equation}
where  $\Pi(\mu,\nu)=\{\G \in \proba{\Omega \times \Omega} | {\pi_1}_\# \G=\mu, {\pi_2}_{\#} \G=\nu\}$ and $\pi_1$ and $\pi_2$ denote the two marginal projections of 
$\Omega \times \Omega$ to $\Omega$. The solutions $\Gzero$ are called optimal transport plans or optimal couplings~\cite{villani2009optimal}.

For $q \geq 1$, the $q$-Wasserstein distance between $\mu$ and $\nu$ in $\mathcal{P}(\Omega)$ is defined as
\begin{equation}
\small
 W_q(\mu, \nu) = \bigg(  \inf_{\G \in \Pi(\mu,\nu)}  \int_{\Omega\times\Omega} c(\x_1,\x_2)^q d\G(\x_1,\x_2) \bigg)^{1/q},
\end{equation}
which is the minimum cost induced by the optimal transportation plan. 
In our method, we use the $1$-Wasserstein distance, also called the earth mover's distance (EMD).

\subsection{Learning with Sliced Wasserstein Discrepancy}
\label{sec:wd}
In this work, we propose to apply $1$-Wasserstein distance to the domain adaptation framework described in Section~\ref{sec:setting}. We utilize the geometrically meaningful $1$-Wasserstein distance as the discrepancy measure in step 2 and step 3 in the aforementioned framework.
In practice, we consider the discrete version of classifiers' logits $p_1(\mathbf{y}|\mathbf{x})$ and $p_2(\mathbf{y}|\mathbf{x})$. Computing $W_1(p_1, p_2)$ requires obtaining the optimal transport coupling $\Gzero$ by solving a linear programming problem~\cite{kantorovitch1958translocation}, which is not efficient.
Although various optimization approaches have been proposed in the past~\cite{cuturi2013sinkhorn, frogner2015learning}, it is unclear how we can directly optimize $W_1(p_1, p_2)$ in an end-to-end trainable fashion efficiently.

To take advantage of the best of both worlds -- to align distributions of source and target by utilizing the task-specific decision boundaries and to incorporate the Wasserstein discrepancy, which has well-behaved energy landscape for stochastic gradient descent training, we propose to integrate $W_1(p_1, p_2)$ into our framework by using the sliced Wasserstein discrepancy, a 1-D variational formulation of the $1$-Wasserstein distance between the outputs $p_1$ and $p_2$ of the classifiers along the radial projections. 

Motivated by~\cite{rabin2011wasserstein} which defines a sliced barycenter of discrete measures, we define the sliced $1$-Wasserstein discrepancy (SWD) as
\begin{equation}
\small
SW\hspace{-0.7mm}D (\mu, \nu) = \int_{\mathcal{S}^{d-1}} W_1( {\mathcal{R}_\theta} \mu, {\mathcal{R}_\theta} \nu) d\theta,
\end{equation}
where ${\mathcal{R}_\theta}$ denotes a one-dimensional linear projection operation on the probability measure $\mu$ or $\nu$, and $\theta$ is a uniform measure on the unit sphere $\mathcal{S}^{d-1}$ in $\mathbb{R}^d$ such that $\int_{\mathcal{S}^{d-1}} d \theta = 1$. In this manner, computing the sliced Wasserstein discrepancy is equivalent to solving several one-dimensional optimal transport problems which have closed-form solutions~\cite{rabin2011wasserstein}.

\input{Tables/alg1.tex}

Specifically, let $\alpha$ and $\beta$ be the permutations that order the $N$ one-dimensional linear projections of $N$ samples such that $\forall \: 0 \leq i < N-1, {{\mathcal{R}_\theta} \mu}_{\alpha(i)} \leq {{\mathcal{R}_\theta} \mu}_{\alpha(i+1)}$ and ${{\mathcal{R}_\theta} \nu}_{\beta(i)} \leq {{\mathcal{R}_\theta} \nu}_{\beta(i+1)}$, then the optimal coupling $\Gzero$ that minimizes such one-dimensional Wasserstein distance is simply assign ${{\mathcal{R}_\theta} \mu}_{\alpha(i)}$ to ${{\mathcal{R}_\theta} \nu}_{\beta(i)}$ using a sorting algorithm.
For discrete probability measures, our SWD can be written as:
 \begin{equation}
 \small
SW\hspace{-0.7mm}D (\mu, \nu) = \sum\limits_{m=1}^{M} \sum\limits_{i=1}^{N}  c({{\mathcal{R}_{\theta_m}} \mu}_{\alpha(i)}, {{\mathcal{R}_{\theta_m}} \nu}_{\beta(i)})
 \end{equation}
 for $M$ randomly sampled $\theta$ and quadratic loss for $c$ unless otherwise mentioned.
Our proposed SWD is essentially a variational version of original Wasserstein distance but at a fraction of its computational cost~\cite{bonneel2015sliced}. More importantly, the SWD is differentiable due to the close-form characteristic, so we can focus on using optimal transport as a reliable fidelity measure to guide the optimization of feature generator and classifiers. 
We summarize our framework in Algorithm~\ref{alg:swd} and illustrate the SWD computation in Figure~\ref{fig:pipeline}.

\begin{figure*}
\begin{center}
\includegraphics[width=0.26\linewidth]{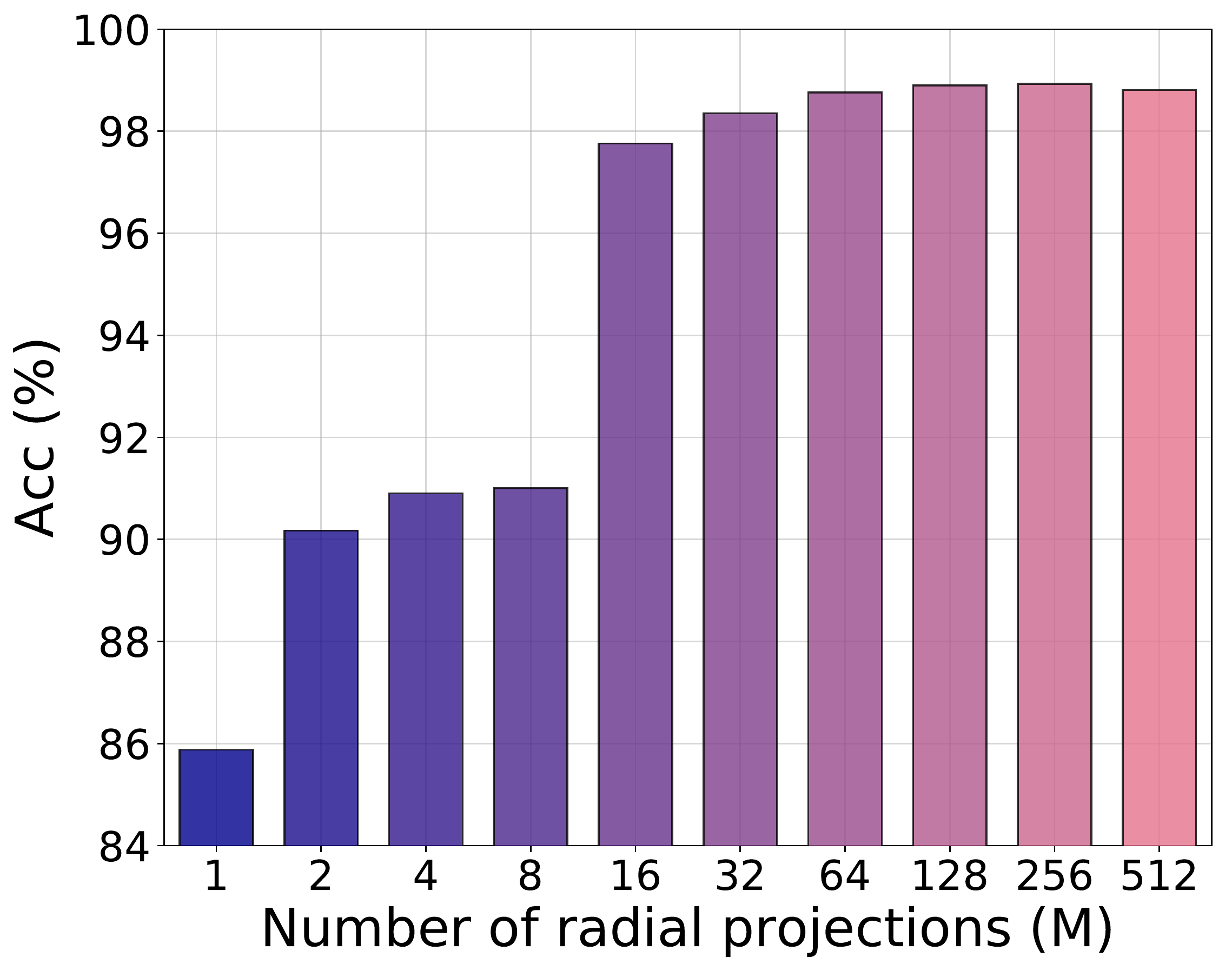}
\includegraphics[width=0.26\linewidth]{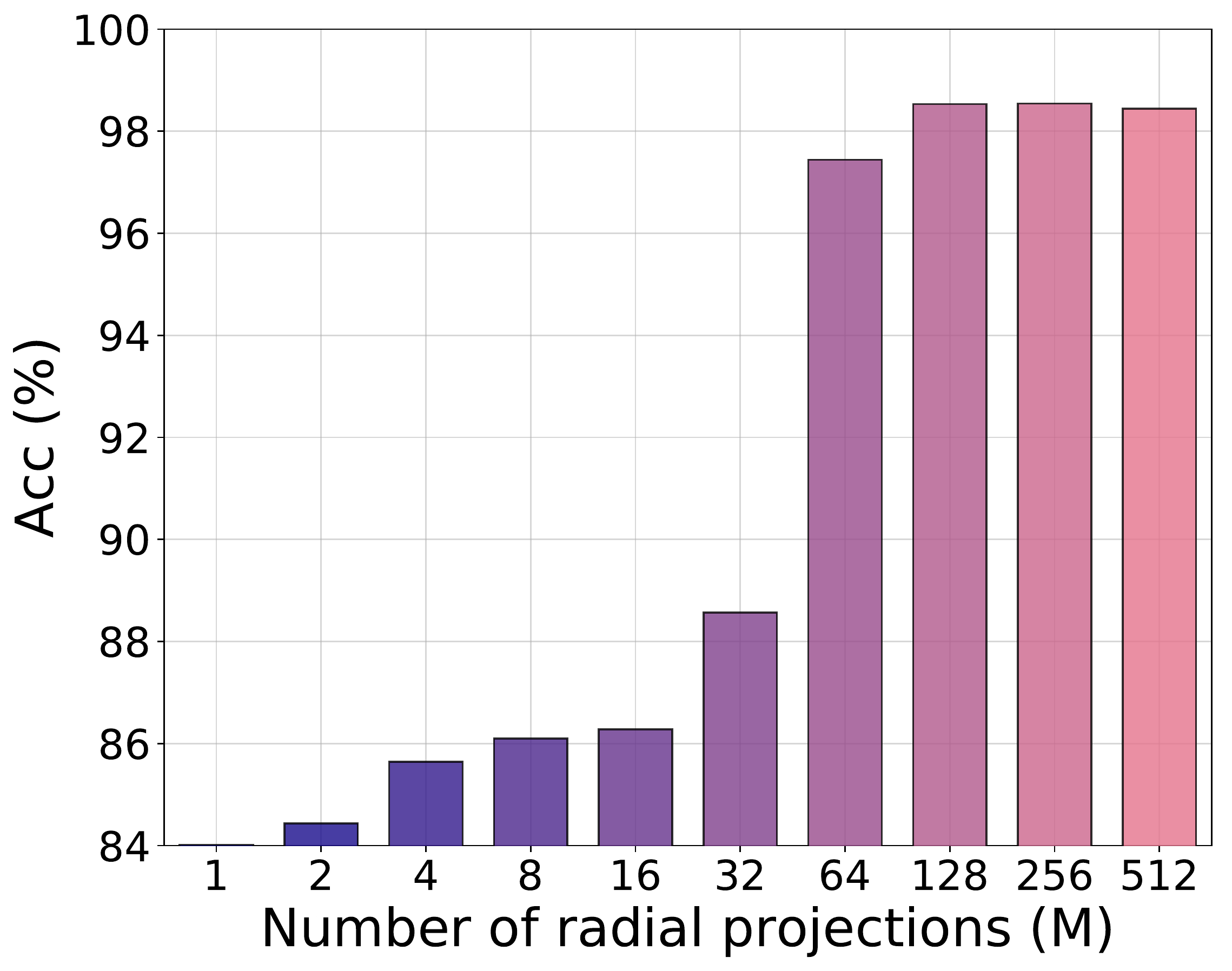}
\includegraphics[width=0.23\linewidth]{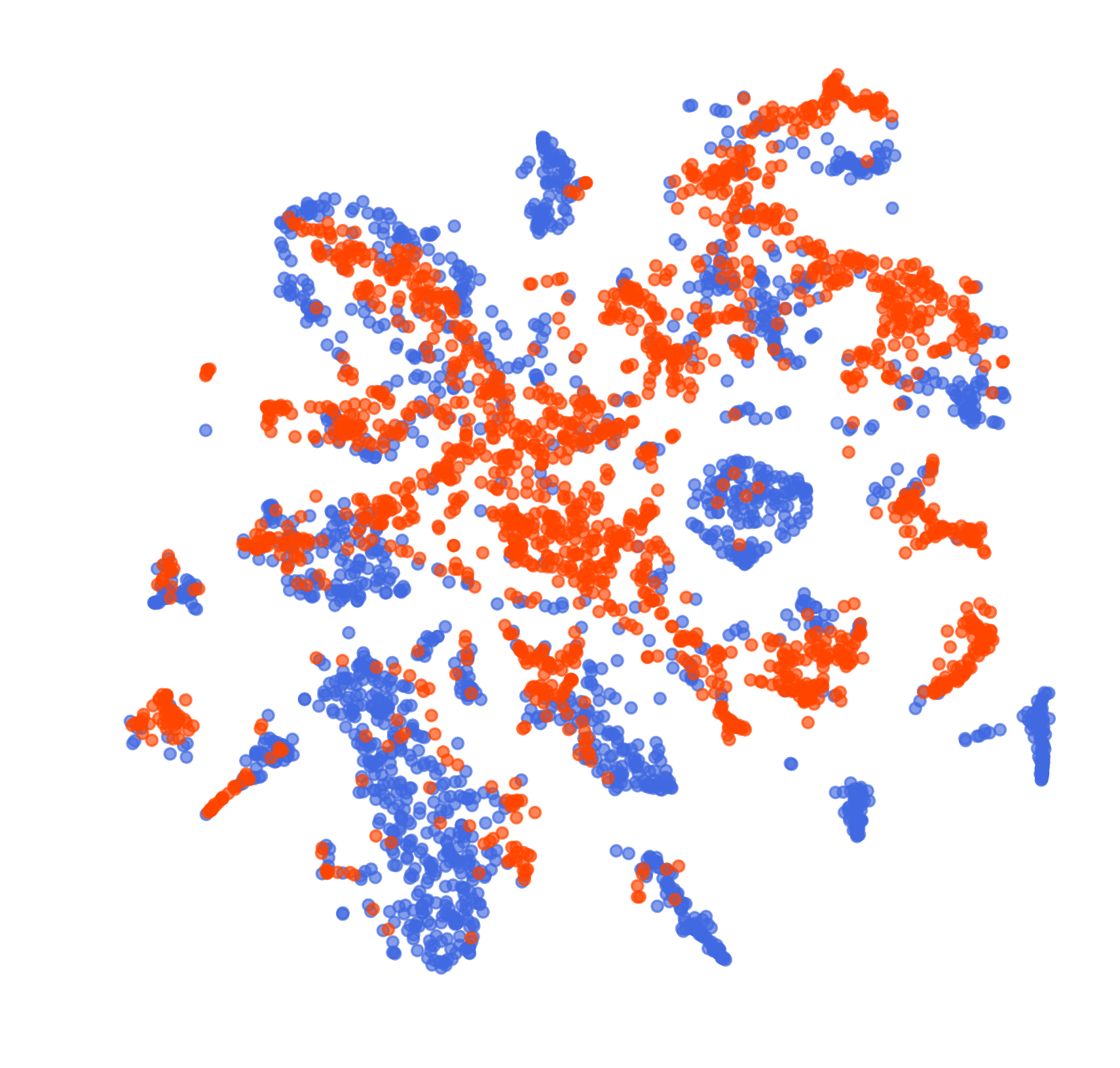}
\includegraphics[width=0.23\linewidth]{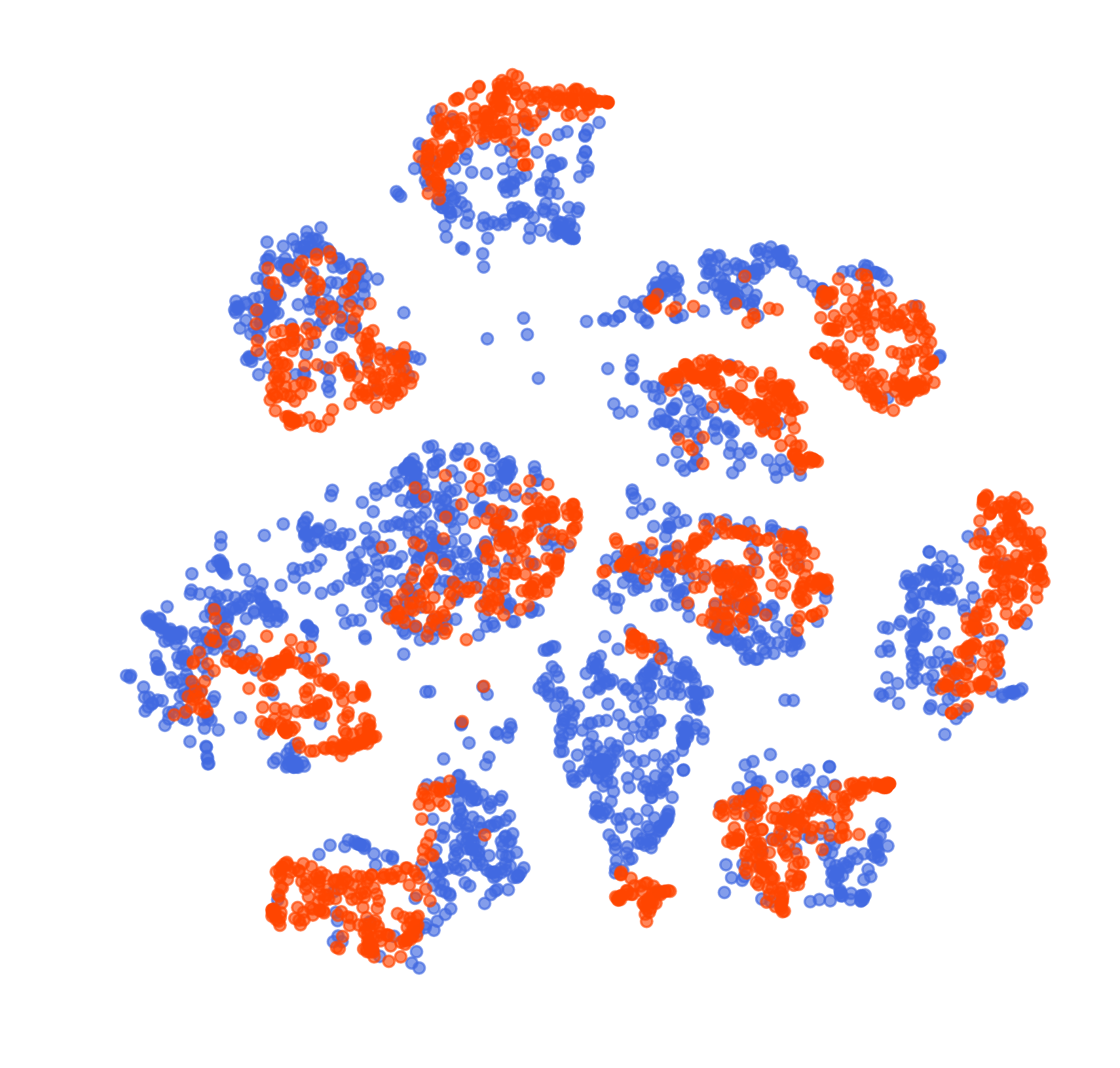}
\end{center}
\vspace{-3mm}
\small \hspace{10mm} (a) SVHN to MNIST  \hspace{17mm} (b) SYNSIG to GTSRB \hspace{17mm} (c) Source only \hspace{13.5mm} (d) SWD adapted (ours)
\vspace{-2mm}
   \caption{The effect of number of radial projections $M$ to accuracy on (a) SVHN to MNIST adaptation, and (b) SYNSIG to GTSRB adaptation. $M=128$ is sufficient for stable optimization and good accuracies. T-SNE~\cite{maaten2008visualizing} visualization of features obtained from SVHN to MNIST experiment by (c) source domain only, and (d) SWD adaptation. Blue and red points denote the source and target samples, respectively. Our method generates much more discriminative feature representation compared to source only setting.}
\label{fig:projection}
\vspace{-2mm}
\end{figure*}

\section{Experiments}
In principle, our method can be applied to any domain adaptation tasks and does not require any similarity assumptions in input or output space. We perform extensive evaluation of the proposed method on digit and sign recognition, image classification,  semantic segmentation, and object detection tasks.

\subsection{Digit and Sign Recognition}
In this experiment, we evaluate our method using five standard benchmark datasets: Street View House Numbers (SVHN)~\cite{netzer2011reading}, MNIST~\cite{lecun1998gradient}, USPS~\cite{hull1994database}, Synthetic Traffic Signs (SYNSIG)~\cite{moiseev2013evaluation}, and German Traffic Sign Recognition Benchmark (GTSRB)~\cite{stallkamp2011german} datasets.
For each domain shift pair, we use the exact CNN architecture provided by Saito \etal~\cite{saito2017maximum}. We use Adam~\cite{kingma2014adam} solver with mini-batch size of 128 in all experiments. Gradient reversal layer (GRL)~\cite{ganin2014unsupervised} is used for training the networks so we do not need to control the update frequency between the generator and classifiers. The hyper-parameter particular to our method is the number of radial projections $M$. We varied the value of $M$ in our experiment and detailed the sensitivity to the hyper-parameter in Figure~\ref{fig:projection}(a) and~\ref{fig:projection}(b).

\vspace{2mm}
\noindent \textbf{SVHN $\rightarrow$ MNIST}
We first examine the adaptation from real-world house numbers obtained from Google Street View images~\cite{netzer2011reading} to handwritten digits~\cite{lecun1998gradient}. The two domains demonstrate distinct distributions because images from SVHN dataset contain cluttered background from streets and cropped digits near the image boundaries.
We use the standard training set as training samples, and testing set as test samples both for source and target domains. The feature generator contains three 5$\times$5 conv layers with stride two 3$\times$3 max pooling placed after the first two conv layers. For classifiers, we use 3-layered fully-connected networks.

\vspace{2mm}
\noindent \textbf{SYNSIG $\rightarrow$ GTSRB}
In this setting, we evaluate the adaptation ability from synthetic images SYNSIG to real images GTSRB. We randomly selected 31367 samples for target training and evaluated the accuracy on the rest. The feature generator contains three 5$\times$5 conv layers with stride two 2$\times$2 max pooling placed after each conv layer. For classifiers, we use 2-layered fully-connected networks. The performance is evaluated for 43 common classes between the two domains.

\vspace{2mm}
\noindent \textbf{MNIST $\leftrightarrow$ USPS}
For the two-way domain shift experiment we also follow the protocols provided by~\cite{saito2017maximum} that we use the standard training set as training samples, and testing set as test samples both for source and target domains. The feature generator contains two 5$\times$5 conv layers with stride two 2$\times$2 max pooling placed after the each conv layer. For classifiers, we use 3-layered fully-connected networks.

\vspace{2mm}
\noindent \textbf{Results}
Table~\ref{tab:digits} lists the accuracies for the target samples by four different domain shifts. We observed that our SWD method outperforms competing approaches in all settings. The proposed method also outperforms the direct comparable method MCD~\cite{saito2017maximum} by a large margin -- absolute accuracy improvement of 2.8\% on average across the four settings. Figure~\ref{fig:projection}(a) and~\ref{fig:projection}(b) show the ablation study on the sensitivity to the number of radial projections $M$. In our experiment we empirically found that $M=128$ works well in all cases. We also visualized learned features in Figure~\ref{fig:projection}(c) and~\ref{fig:projection}(d). Our method generates much more discriminative feature representation compared to source only setting.

\input{Tables/digits.tex}

\input{Tables/visda2017.tex}

It is interesting to see that the task-specific discrepancy-aware methods such as MCD~\cite{saito2017maximum}, DeepJDOT~\cite{damodaran2018deepjdot}, and the proposed SWD are the current leading approaches for the tasks being addressed here. This demonstrates the importance of utilizing the task-specific decision boundaries (discrepancy) to guide the process of transfer learning instead of simply matching the distributions between the source and target domains in pixel, feature, or output space in most of the other distribution matching approaches. In particular, the adversarial training based methods require a separate generator and multiple discriminators that are oftentimes larger than the main task network itself. For example, the method in~\cite{hoffman2017cycada} uses a 10-layer generator, a 6-layer image level discriminator, and a 3-layer feature level discriminator while the main task network is a 4-layer network. Besides, the auxiliary discriminators are discarded after the training is completed.

Furthermore, the main distinction between the proposed method and the DeepJDOT~\cite{damodaran2018deepjdot} approach is that the DeepJDOT requires a multi-staged training process -- it trains a CNN and solves a linear programming task iteratively. The DeepJDOT also assumes the true optimal transport coupling between every pair of samples in a mini-batch converges when propagating pseudo labels from the source domain to target domain, which is often not the case in practice. This emphasizes the importance of choosing a geometrically meaningful discrepancy measure that makes no assumptions on the optimal transport coupling in the label space and is end-to-end trainable, optimizing over one discrepancy loss instead of solving multiple losses independently.

We note that the method in~\cite{shu2018dirt} obtained 99.4\% on SVHN to MNIST adaptation task with various engineering efforts such as using instance normalization, adding Gaussian noise, and leveraging a much deeper 18-layer network. This clustering assumption-based approach achieves performance on-par with ours, and we leave this architectural search for future exploration.

\subsection{VisDA Image Classification}
Next, we evaluate the proposed method on image classification task. We use the VisDA dataset~\cite{visda2017}, which is designed to assess the domain adaptation capability from synthetic to real images across 12 object classes. The source domain contains 152,397 synthetic images generated by rendering 3D CAD models from different angles and under different lighting conditions. For target domain we use the validation set collected from MSCOCO~\cite{lin2014microsoft} and it consists of 55,388 real images.
Following the protocol in~\cite{saito2017maximum}, we evaluate our method by fine-tuning ImageNet~\cite{deng2009imagenet} pre-trained ResNet-101~\cite{he2016deep} model. The ResNet model except the last fully-connected layer was used as our feature generator and randomly initialized three-layered fully-connected networks were used as our classifiers. We used Adam solver with mini-batch size of 32 in the experiment. The number of radial projections $M$ is set to 128. We apply horizontal flipping of input images during training as the only data augmentation. Learning rate was set to value of 10$^{-6}$ throughout the training. GRL~\cite{ganin2014unsupervised} is used for training the network so we do not need to control the update frequency between the generator and classifiers.

\input{Tables/gta2city.tex}
\input{Tables/synthia2city.tex}
\begin{figure*}[h]
\begin{center}
\includegraphics[width=0.8\linewidth]{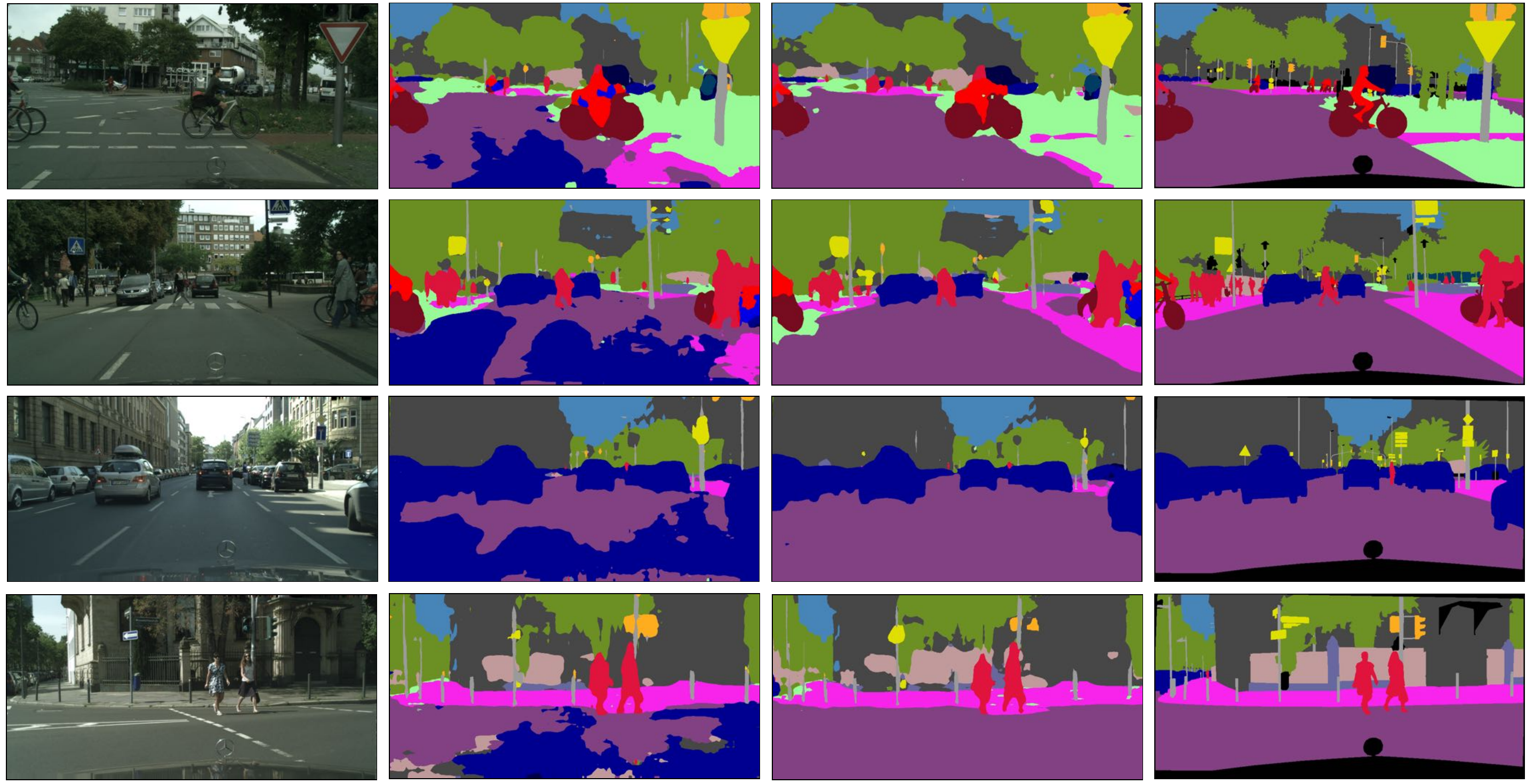}
\end{center}
\vspace{-6mm}
   \caption{Qualitative semantic segmentation results on adaptation from GTA5 to Cityscapes. From left to right: input, source only model, our method, and ground truth. Our method produces cleaner predictions and less confusion between challenging classes such as road, car, sidewalk, and vegetation.}
\label{fig:segmentation}
\vspace{-3mm}
\end{figure*}

\vspace{2mm}
\noindent \textbf{Results}
Table~\ref{tab:visda2017} lists the results that are based on the same evaluation protocol\footnote{Method in~\cite{french2017self} involves various data augmentation including scaling, cropping, flipping, rotation, brightness and color space perturbation, etc. The authors reported mean accuracy of 74.2$\%$ in the minimal augmentation setting using a much deeper ResNet-152 backbone.}. We can see that the task-specific discrepancy-aware methods MCD~\cite{saito2017maximum} and our SWD method perform better than the source only model in all object classes, while pure distribution matching based methods perform worse than the source only model in some categories.
Our method also outperforms the direct comparable method MCD~\cite{saito2017maximum} by a large margin. We emphasize that the main difference between MCD and our method is the choice of discrepancy loss. This validated the effectiveness of the proposed sliced Wasserstein discrepancy in this challenging synthetic to real image adaption task.

\subsection{Semantic Segmentation}
Unlike image classification task, obtaining ground truth label for each pixel in an image requires a lot more amount of human labor. Here we extend our framework to perform domain adaptation for the semantic segmentation task.

\vspace{2mm}
\noindent \textbf{Datasets}
In this experiment we used three benchmark datasets: GTA5~\cite{richter2016playing}, Synthia~\cite{ros2016synthia}, and Cityscapes~\cite{cordts2016cityscapes}. All three datasets contain dense pixel-level semantic annotations that are compatible with one another. GTA5 contains 24966 vehicle-egocentric images synthesized from a photorealistic open-world computer game Grand Theft Auto V. Synthia consists of 9400 images generated by rendering a virtual city created with the Unity engine. Frames in Synthia are acquired from multiple camera viewpoints -- up to eight views per location that are not necessary vehicle-egocentric. Cityscapes has 2975 training images, 500 validation images, and 1525 test images with dense pixel-level annotation of urban street scenes in Germany and neighboring countries. During training, the synthetic GTA5 or Synthia is used as source domain and the real Cityscapes is used as target domain. We used the standard training split for training and validation split for evaluation purpose.

\input{Tables/detection.tex}

\vspace{2mm}
\noindent \textbf{Implementation details}
In an effort to demonstrate the effectiveness of the proposed method and to decouple the performance gain due to architectural search, we adopted the commonly used VGG-16~\cite{simonyan2014very} and ResNet-101~\cite{he2016deep} models for our feature generator. For classifiers we used the decoder in PSPNet~\cite{zhao2017pyramid} for its simplicity of implementation. Based on this straightforward design choice, our segmentation model achieves mean intersection-over-union (mIoU) of 60.5$\%$ for VGG-16 backbone and 64.1$\%$ for ResNet-101 backbone when trained on the Cityscapes training set and evaluated on the Cityscapes validation set for 19 compatible classes, which match the same oracle performance reported in the recent literature~\cite{tsai2018learning, wu2018dcan}.

We used Momentum SGD solver with a fixed momentum of 0.9 and weight decay of 0.0001 in all experiments. Learning rate was set to value of 0.0001 for GTA5 to Cityscapes setting and 0.001 for Synthia to Cityscapes setting. During training, we randomly sampled a single image from both source and target domains for each mini-batch optimization. All images are resized to 1024$\times$512 resolution. No data augmentation is used (such as flipping, cropping, scaling, and multi-scale ensemble) to minimize the performance gain due to the engineering efforts and to ensure the reproducibility.
Since the sliced Wasserstein discrepancy is computed at every pixel of an image in a mini-batch fashion, we empirically found that the number of radial projections $M=8$ is sufficient to produce good results.

\vspace{2mm}
\noindent \textbf{Results}
We use the evaluation protocol released along with VisDA challenge~\cite{visda2017} to calculate the PASCAL VOC intersection-over-union (IoU). We show quantitative and qualitative results of adapting GTA5 to Cityscapes in Table~\ref{tb:gta2city} and Figure~\ref{fig:segmentation}, respectively. We can see clear improvement from models trained on source domain only to models trained with the proposed SWD method for both VGG-16 and ResNet-101 backbones. Also, our method consistently outperforms other recent approaches that utilize generative adversarial networks~\cite{tsai2018learning, hoffman2017cycada} and the style transfer based technique~\cite{wu2018dcan}.

Table~\ref{tb:synthia2city} shows the results of adapting Synthia to Cityscapes. 
The domain shift is even more significant between these two datasets because images from Synthia are not only generated by a rendering engine but also contain multiple viewpoints that are not necessary vehicle-egocentric. Our method shows consistent improvement over other approaches and generalizes well with such dramatic viewpoint shift.

Note that most of the competing approaches are specifically designed only for semantic segmentation tasks and they often require the assumption of input space similarity, output space similarity, or geometric constrains. For instance, the method in~\cite{zou2018unsupervised} incorporated spatial priors by assuming sky is likely to appear at the top and road is likely to appear at the bottom of an image, etc. The frequencies of ground truth labels per pixel are computed from GTA5 dataset and are then multiplied with the softmax output of the segmentation network. However, it is unclear how to generalize this prior when large viewpoint differences present between the source and target domains, such as adaptation from Synthia to Cityscapes. Our method does not require any prior assumptions of the characteristics of the task of interest and nevertheless achieves better performance.

\subsection{Object Detection}
To demonstrate if our method generalizes to other tasks as well, we extend it to object detection task. We use the recent released VisDA 2018 dataset~\cite{peng2018syn2real}, which contains source images generated by rendering 3D CAD models and target images collected from MSCOCO~\cite{lin2014microsoft}. This dataset is especially challenging due to uncalibrated object scales and positions between the synthetic and real images.

We use a standard off-the-shelf Single Shot Detector (SSD)~\cite{liu2016ssd} with Inception-V2~\cite{szegedy2016rethinking} backbone without any architectural modifications or heuristic assumptions. The model predicts class labels, locations and size shifts for a total of 1.9k possible anchor boxes. The feature generator in this case is the backbone network pre-trained on ImageNet and the classifiers comprise of all the additional layers which are present after the backbone network. We employ the proposed sliced Wasserstein discrepancy to both classification and bounding box regression outputs to the existing loss functions in SSD. No other modifications are made to the network. We also implement MCD~\cite{saito2017maximum} method with the exact network architecture for baseline comparison. All networks are optimized with Momentum SGD solver with a fixed momentum of 0.9, mini-batch size of 16, and weight decay of 0.0001. Learning rate is set to value of 0.0001. The number of radial projections $M$ is set to 128. We apply random cropping and flipping to all network training.

\vspace{2mm}
\noindent \textbf{Results}
We report mean average precision (mAP) at 0.5 IoU in Table~\ref{tb:detection}. These results show that even with large domain shift in image realism, object scales, and relative object positions, our method is able to improve the performance by a large margin compared to models trained on source image only. Our method also outperforms the direct comparable method MCD~\cite{saito2017maximum} by a significant 25\% relatively.

\section{Conclusion}
In this paper, we have developed a new unsupervised domain adaptation approach, which aligns distributions by measuring sliced Wasserstein discrepancy between task-specific classifiers. The connection to Wasserstein metric paves the way to make better use of its geometrically meaningful embeddings in an efficient fashion, which in the past has primarily been restricted to obtaining one-to-one mapping in label space. Our method is generic and achieves superior results across digit and sign recognition, image classification, semantic segmentation, and object detection tasks. Future work includes extension of our approach to domain randomization~\cite{sundermeyer2018implicit}, open set adaptation~\cite{saito2018open}, and zero-shot domain adaptation~\cite{peng2018zero}.

{\small
\bibliographystyle{ieee}
\bibliography{egbib}
}

\section*{Supplementary Material}
\vspace{-1mm}
We perform a toy experiment on the inter twinning moons 2D dataset~\cite{pedregosa2011scikit} to provide analysis on the learned decision boundaries as shown in Figure~\ref{fig:toy} below.
For the source samples, we generate an upper moon (blue points) and a lower moon (red points), labeled 0 and 1, respectively. The target samples are generated from the same distribution as the source samples but with added domain shifts by rotation and translation. The model consists of a 3-layered fully-connected network for a feature generator and 3-layered fully-connected networks for classifiers.

After convergence, the source only model (Figure~\ref{fig:toy}{\color{red}(a)}) classifies the source samples perfectly but does not generalize well to the target samples in region 1 and region 2.
The MCD approach (Figure~\ref{fig:toy}{\color{red}(b)}) is able to adapt its decision boundary correctly in region 1 but not in region 2.
The proposed SWD method (Figure~\ref{fig:toy}{\color{red}(c)}) adapts to the target samples nicely and draws a correct decision boundary in all regions.

\vspace{-1mm}
\begin{figure}[h]
\hspace{0.05mm} \includegraphics[width=0.38\linewidth]{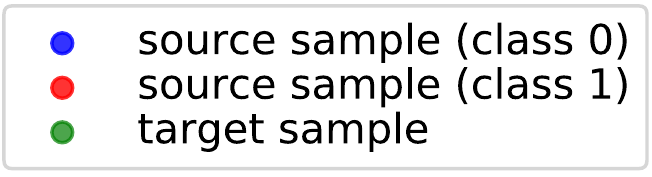}
\vspace{-7mm}
\end{figure}

\begin{figure}[h]
\begin{center}
\includegraphics[width=0.32\linewidth]{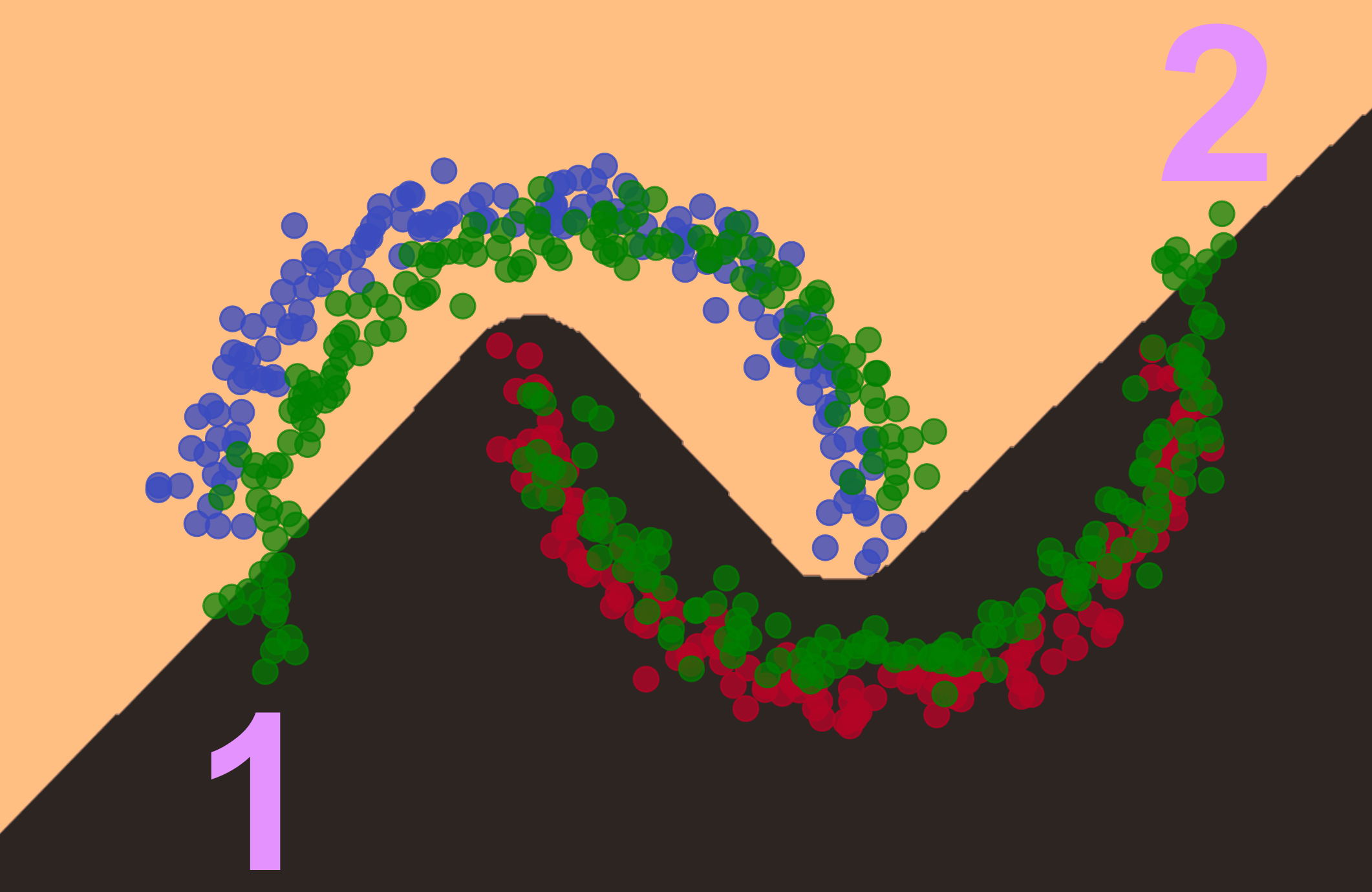}
\includegraphics[width=0.32\linewidth]{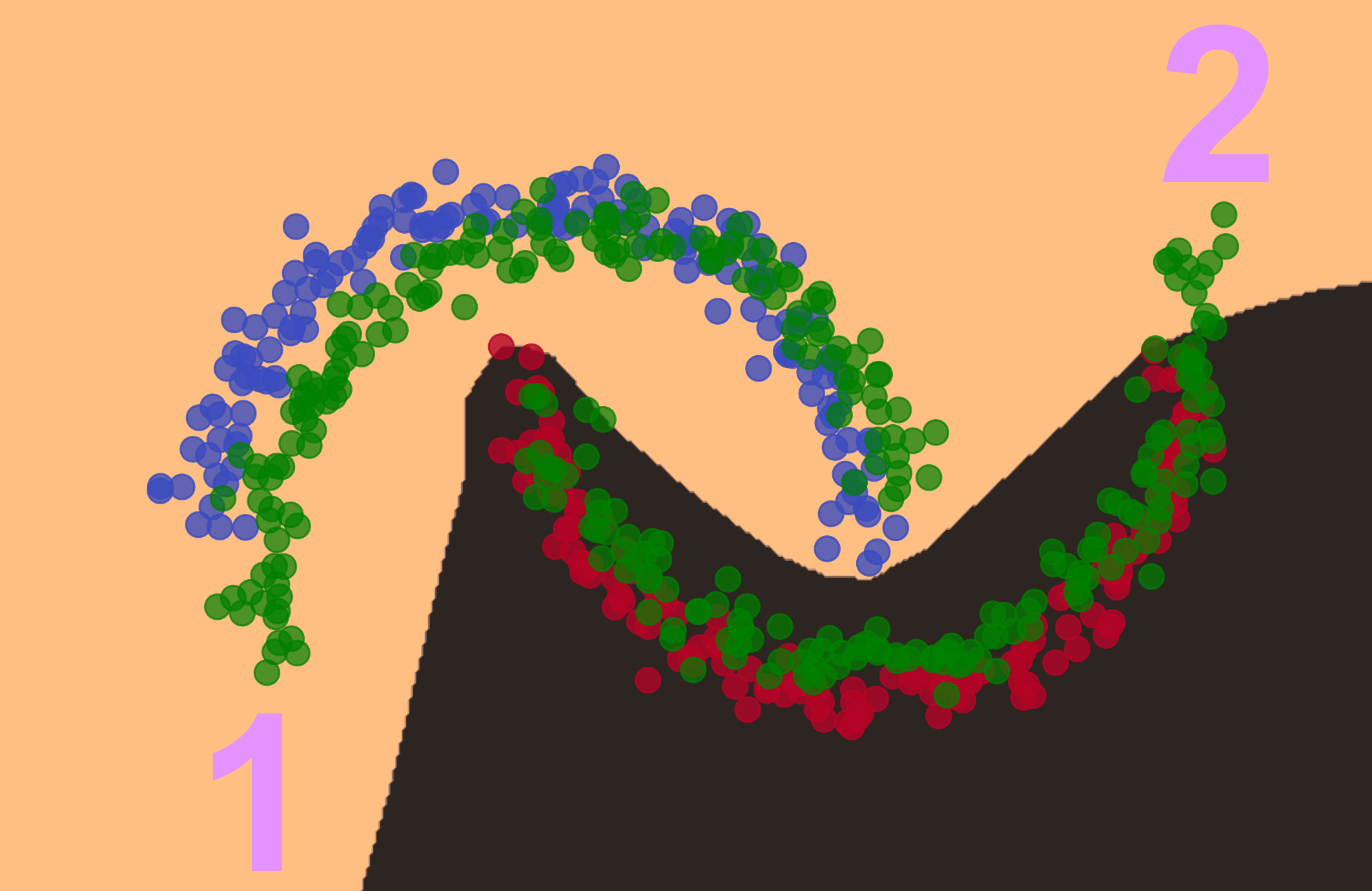}
\includegraphics[width=0.32\linewidth]{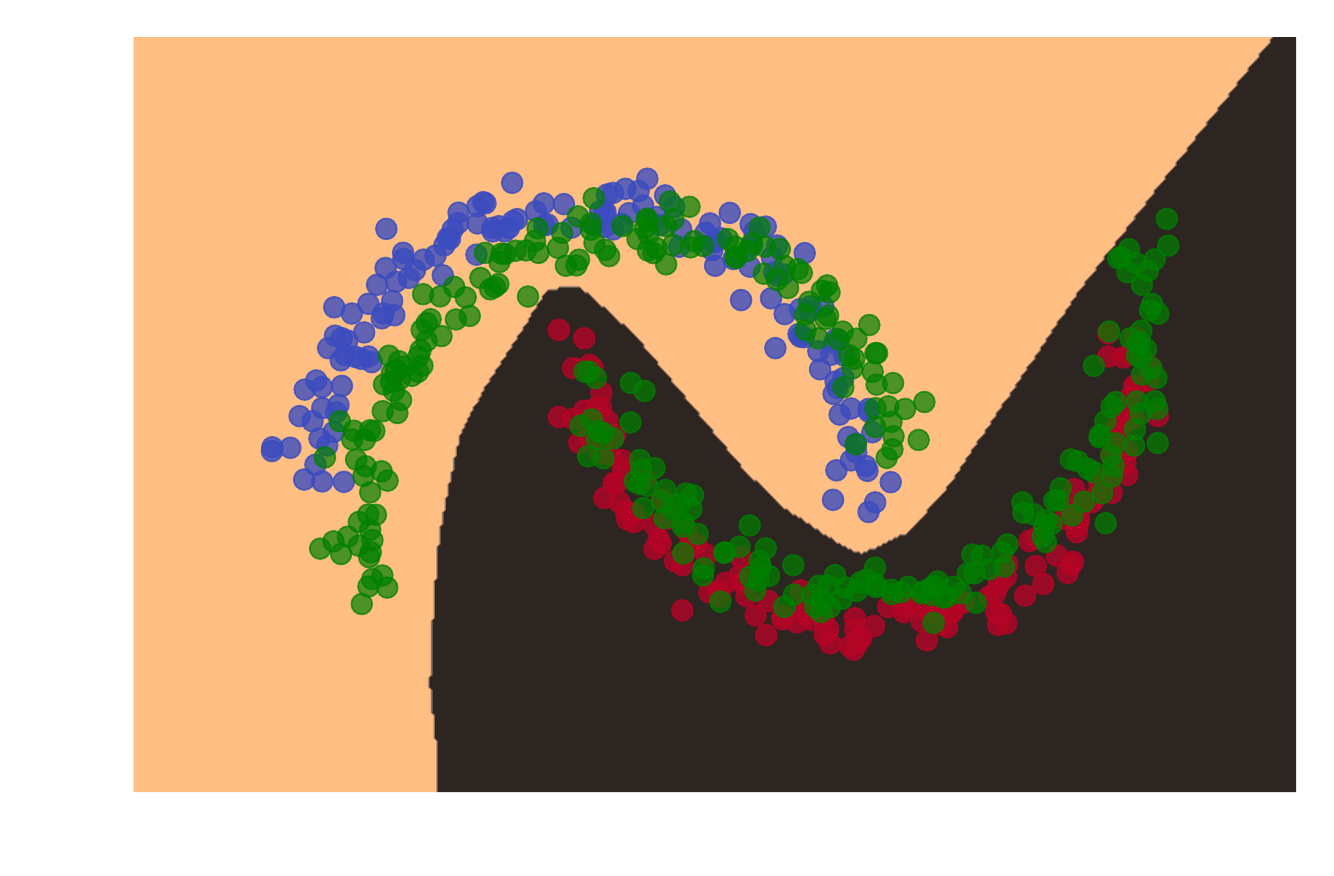}
\end{center}
\vspace{-4mm}
\small \hspace{3mm} (a) Source only \hspace{10mm} (b) MCD  \hspace{12mm} (c) SWD (ours)
\vspace{-2mm}
   \caption{Comparison of three decision boundaries on a toy example. Blue and red points indicate the source samples of class 0 and 1, respectively. Green points are the target samples generated from the same distribution as the source samples but with domain shifts by rotation and translation. The orange and black regions are classified as class 0 and 1, respectively, after convergence.}
\label{fig:toy}
\end{figure}

\end{document}

%% file: Tables/alg1.tex
\begin{algorithm}[t!]
\footnotesize
\caption{Sliced Wasserstein Discrepancy for Unsupervised Domain Adaptation}
\label{alg:swd}
\begin{algorithmic}
\REQUIRE Labeled source set \{$X_{s}, Y_{s}$\}, unlabeled target set $X_{t}$, number of random projections $M$, and randomly initialized feature generator $G$ and classifiers $C_1, C_2$.
\WHILE{$G$, $C_1$, and $C_2$ have not converged}
\STATE Step 1: Train $G$, $C_1$, and $C_2$ on labeled source set:
\STATE
\setlength{\leftskip}{8.8mm}
\vspace{-2.5mm}
 \begin{equation*}
   \min_{G, C_1, C_2} \mathcal{L}_s (X_{s},Y_{s})
 \end{equation*}
\setlength{\leftskip}{0pt}
 \vspace{-2mm}
\STATE Step 2: Train $C_1$ and $C_2$ to maximize sliced Wasserstein discrepancy
\STATE
\setlength{\leftskip}{8.8mm}
 \vspace{-3.5mm}
on unlabeled target set:
\STATE Obtain classifiers' output ${p_1}$ and ${p_2}$ on target samples
\STATE Sample $\{\theta_1,...,\theta_M\}$ from $\mathcal{S}^{d-1}$ in $\mathbb{R}^d$
\STATE Sort ${\mathcal{R}_{\theta_m}} {p_1}$ such that ${{\mathcal{R}_{\theta_m}} {p_1}_{(j)}} \leq {{\mathcal{R}_{\theta_m}} {p_1}_{(j+1)}}$ 
\STATE Sort ${\mathcal{R}_{\theta_m}} {p_2}$ such that ${{\mathcal{R}_{\theta_m}} {p_2}_{(j)}} \leq {{\mathcal{R}_{\theta_m}} {p_2}_{(j+1)}}$
 \vspace{-1mm}
 \begin{equation*}
  \min_{C_1, C_2} \mathcal{L}_s (X_{s},Y_{s}) - \mathcal{L}_{\text{DIS}}(X_{t})
 \end{equation*}
 \vspace{-3mm}
\STATE where $\mathcal{L}_{\text{DIS}}(X_{t}) = \sum\limits_{m} \sum\limits_{j}  c({{\mathcal{R}_{\theta_m}} {p_1}_{(j)}}, {{\mathcal{R}_{\theta_m}} {p_2}_{(j)}})$

\setlength{\leftskip}{0pt}
\STATE Step 3: Train $G$ to minimize the same sliced Wasserstein discrepancy
\STATE
\setlength{\leftskip}{8.8mm}
\vspace{-3.5mm}
on unlabeled target set:
 \vspace{-1.5mm}
 \begin{equation*}
  \min_{G} \mathcal{L}_{\text{DIS}}(X_{t})
 \end{equation*}
\setlength{\leftskip}{0pt}
 \vspace{-5mm}
\ENDWHILE
\end{algorithmic}
\end{algorithm}

%% file: Tables/digits.tex
\begin{table}
\footnotesize{
\setlength{\tabcolsep}{5.1pt}
\def\arraystretch{1.1}
\begin{center}
\scalebox{0.92}{
\begin{tabular}{l | c | c | c | c}
    \toprule
    		\vspace{-0.5mm}
             & \scriptsize{SVHN} & \scriptsize{SYNSIG} & \scriptsize{MNIST} & \scriptsize{USPS} \\
             \vspace{-0.5mm}
Method & $\downarrow$  & $\downarrow$  & $\downarrow$  & $\downarrow$  \\
             & \scriptsize{MNIST} & \scriptsize{GTSRB} & \scriptsize{USPS} & \scriptsize{MNIST}  \\
	\midrule 
	\midrule
	Source only &67.1&85.1&79.4&63.4 \\
	MMD~\cite{long2015learning} & 71.1&91.1&81.1&-\\
	DANN~\cite{ganin2014unsupervised} & 71.1&88.7&85.1&73.0 $\pm$ 0.2 \\
	DSN~\cite{bousmalis2016domain} & 82.7&93.1&-&-\\
	ADDA~\cite{tzeng2017adversarial} &76.0 $\pm$ 1.8&-&-&90.1 $\pm$ 0.8 \\
  	CoGAN~\cite{liu2016coupled}&-&-&-&89.1 $\pm$ 0.8\\
  	PixelDA~\cite{bousmalis2017unsupervised}&-&-&95.9&-\\
	ASSC~\cite{haeusser2017associative} &95.7 $\pm$ 1.5&82.8 $\pm$ 1.3&-&-\\
	UNIT~\cite{liu2017unsupervised} & 90.5 & - & 96.0  & 93.6\\
	CyCADA~\cite{hoffman2017cycada} & 90.4 $\pm$ 0.4 & - & 95.6 $\pm$ 0.2 & 96.5 $\pm$ 0.1 \\
	I2I Adapt~\cite{murez2018image} & 92.1 & - & 95.1 & 92.2 \\
	GenToAdapt~\cite{sankaranarayanan2017generate} & 92.4 $\pm$ 0.9 & - & 95.3 $\pm$ 0.7 & 90.8 $\pm$ 1.3 \\
 	MCD~\cite{saito2017maximum}&96.2 $\pm$ 0.4& 94.4 $\pm$ 0.3& 96.5 $\pm$ 0.3& 94.1 $\pm$ 0.3\\
	DeepJDOT~\cite{damodaran2018deepjdot} & 96.7 & - & 95.7 & 96.4 \\
	\midrule
	SWD (ours) & {\bf98.9} $\pm$ 0.1 &   {\bf98.6} $\pm$ 0.3 & {\bf98.1} $\pm$ 0.1 & {\bf97.1} $\pm$ 0.1 \\
    \bottomrule
\end{tabular}}
\end{center}
\vspace{-6mm}
\caption{Results of unsupervised domain adaptation across digit and traffic sign datasets. We repeat each experiment 5 times and report the mean and the standard deviation of the accuracy. Our method significantly outperforms the direct comparable method MCD~\cite{saito2017maximum} and other methods as well.}
\label{tab:digits}}
\vspace{-3mm}
\end{table}

%% file: Tables/visda2017.tex
\begin{table*}[tbp]
\begin{center}
\scalebox{0.7}{
\begin{tabular}{l|cccccccccccc | c}
    \toprule
	Method 		& plane 	& bcycl 	& bus  	& car  	& horse 	& knife 	& mcycl 	& person 	& plant 	& sktbrd 	& train 	& truck 	& mean \\
	\midrule
	\midrule
	Source only 	& 55.1	& 53.3    	& 61.9 	& 59.1 	& 80.6  	& 17.9  	& 79.7   	& 31.2   	& 81.0    	& 26.5     	& 73.5  	& 8.5   	& 52.4 \\
	MMD~\cite{long2015learning} & 87.1 & 63.0      	& 76.5 	& 42.0 	& 90.3  	& 42.9  	& 85.9 & 53.1   	& 49.7  	& 36.3      	& \bf 85.8 	& 20.7  	& 61.1 \\
	DANN~\cite{ganin2014unsupervised} & 81.9      	& 77.7   	& 82.8 	& 44.3 	& 81.2  	& 29.5  	& 65.1   	& 28.6   	& 51.9  	& 54.6     	& 82.8  	& 7.8   	& 57.4 \\
	MCD	~\cite{saito2017maximum} & 87.0	& 60.9	& \bf 83.7	& 64.0	& 88.9	& \bf 79.6	& 84.7	& 76.9	&  \bf 88.6	& 40.3	& 83.0	& 25.8	& 71.9 \\
	\midrule
	SWD (ours) 	 & \bf 90.8 & \bf 82.5 & 81.7 & \bf 70.5 & \bf 91.7 & 69.5 & \bf 86.3 & \bf 77.5 & 87.4 & \bf 63.6 & 85.6 & \bf 29.2 & \bf 76.4 \\
    \bottomrule
  \end{tabular}
  }
  \end{center}
\vspace{-6mm}
\caption{Results of unsupervised domain adaptation on VisDA 2017~\cite{visda2017} image classification track. Accuracies are obtained by fine-tuning ResNet-101 model pre-trained on ImageNet. This task evaluates the adaptation capability from synthetic CAD model images to real-world MSCOCO images. Our method outperforms the direct comparable method MCD~\cite{saito2017maximum} and other methods as well.}
\label{tab:visda2017}
\end{table*}

%% file: Tables/gta2city.tex
\begin{table*}[tbp]
\begin{center}
\scalebox{0.7}{
\begin{tabular}{l|ccccccccccccccccccc | c}
\toprule
Method    & \rot{road} & \rot{sdwk} & \rot{bldng} & \rot{wall} & \rot{fence} & \rot{pole} & \rot{light} & \rot{sign} & \rot{vgttn} & \rot{trrn} & \rot{sky}  & \rot{person} & \rot{rider} & \rot{car}  & \rot{truck} & \rot{bus}  & \rot{train} & \rot{mcycl} & \rot{bcycl} & \rot{mIoU} \\
\midrule
\midrule
Source only (VGG16)   & 25.9 & 10.9 & 50.5 & 3.3  & 12.2 & 25.4 & 28.6 & 13.0 & 78.3 & 7.3  & 63.9 & 52.1 & 7.9 & 66.3 & 5.2  & 7.8  & 0.9 & 13.7 & 0.7 & 24.9 \\
FCN Wld~\cite{hoffman2016fcns} & 70.4 & 32.4 & 62.1  & 14.9 & 5.4   & 10.9 & 14.2  & 2.7  & 79.2  & 21.3 & 64.6 & 44.1   & 4.2   & 70.4 & 8.0     & 7.3  & 0.0     & 3.5   & 0.0  & 27.1   \\
MCD~\cite{saito2017maximum} & 86.4 & 8.5  & 76.1 & 18.6 & 9.7  & 14.9 & 7.8  & 0.6  & \bf 82.8 & \bf 32.7 & 71.4 & 25.2 & 1.1 & 76.3 & 16.1 & 17.1 & 1.4 & 0.2  & 0.0 & 28.8 \\
CDA~\cite{zhang2017curriculum} & 74.9 &22.0 &71.7 &6.0 &11.9 &8.4 &16.3 &11.1 & 75.7 & 13.3 & 66.5 & 38.0 & 9.3 & 55.2 & 18.8 & 18.9 & 0.0 & 16.8 & 14.6 & 28.9 \\
AdaSegNet~\cite{tsai2018learning}  & 87.3 & 29.8 & \bf 78.6 & 21.1 &  18.2 & 22.5 & 21.5 & 11.0 & 79.7 & 29.6 & 71.3 & 46.8 & 6.5 & \bf 80.1 & 23.0 & \bf 26.9 &0.0 & 10.6 & 0.3 & 35.0 \\
CyCADA~\cite{hoffman2017cycada} & 85.2 &	 37.2 &76.5 & 21.8 &15.0 & 23.8 & 22.9 & 	21.5 	& 80.5 & 31.3 &60.7 &50.5 & 9.0 & 76.9 &17.1 &28.2 & \bf 4.5 &9.8	&0.0& 35.4 \\
CBST~\cite{zou2018unsupervised}  & 90.4 & \bf 50.8 & 72.0 & 18.3 & 9.5 & 27.2 & 28.6 & 14.1 & 82.4 & 25.1 & 70.8 & 42.6 & 14.5 & 76.9 & 5.9 & 12.5 & 1.2 & 14.0 & 28.6 & 36.1 \\
DCAN~\cite{wu2018dcan} & 82.3 & 26.7 & 77.4 & \bf 23.7 & 20.5 & 20.4 & \bf 30.3 & 15.9 & 80.9 & 25.4 & 69.5 & 52.6 & 11.1 & 79.6 & \bf 24.9 & 21.2 & 1.30 & 17.0 & 6.70 & 36.2 \\
SWD (ours)  & \bf 91.0 & 35.7 & 78.0 & 21.6 & \bf 21.7 & \bf 31.8 &  30.2 & \bf 25.2 & 80.2 & 23.9 & \bf 74.1 & \bf 53.1 & \bf 15.8 & 79.3 & 22.1 & 26.5 & 1.5 & \bf 17.2 & \bf 30.4 & \bf 39.9 \\ 
\midrule
\midrule
Source only (ResNet101) & 75.8 & 16.8 & 77.2 & 12.5 & 21.0 & 25.5 & 30.1 & 20.1 & 81.3 & 24.6 & 70.3 & 53.8 & 26.4 & 49.9 & 17.2 & 25.9 & \bf 6.5 & 25.3 & \bf 36.0 & 36.6\\
AdaSegNet~\cite{tsai2018learning} & 86.5 & 36.0 & 79.9 & 23.4 & 23.3 & 23.9 & \bf 35.2 & 14.8 & 83.4 & \bf 33.3 & 75.6 & \bf 58.5 & \bf 27.6 & 73.7 & \bf 32.5 & \bf 35.4 & 3.9 & \bf 30.1 & 28.1 & 42.4 \\
SWD (ours) & \bf 92.0 & \bf 46.4 & \bf 82.4 & \bf 24.8 & \bf 24.0 &  \bf 35.1 & 33.4 & \bf 34.2 & \bf 83.6 &  30.4 & \bf 80.9 & 56.9 & 21.9 &  \bf 82.0 & 24.4 & 28.7 & 6.1 & 25.0 & 33.6 & \bf 44.5 \\
\bottomrule
  \end{tabular}
  }
  \end{center}
   \vspace{-6mm}
     \caption{Adaptation results from GTA5 to Cityscapes. We compare our results with other state-of-the-art approaches that are based on the standard VGG-16 or ResNet-101 backbone.}
      \label{tb:gta2city}
\end{table*}

%% file: Tables/synthia2city.tex
\begin{table*}[tbp]
\begin{center}
\scalebox{0.7}{
\begin{tabular}{l|ccccccccccccc|c }
\toprule
Method    & \rot{road} & \rot{sdwk} & \rot{bldng}  & \rot{light} & \rot{sign} & \rot{vgttn} & \rot{sky}  & \rot{person} & \rot{rider} & \rot{car}  & \rot{bus} & \rot{mcycl} & \rot{bcycl} & \rot{mIoU} \\
\midrule
\midrule
Source only (VGG16) & 6.4 & 17.7 & 29.7 & 0.0 & 7.2 & 30.3 &  66.8 &  51.1 & 1.5 &  47.3 & 3.9 & 0.1 & 0.0 & 20.2 \\
FCN Wld~\cite{hoffman2016fcns}  & 11.5 & 19.6 & 30.8 & 0.1 & 11.7 & 42.3 & 68.7 & \bf 51.2 & 3.8 & 54.0 & 3.2 & 0.2 & 0.6 & 22.9  \\
Cross-city~\cite{chen2017no}  & 62.7 & 25.6 & 78.3 & 1.2 & 5.4 & 81.3 & 81.0 & 37.4 & 6.4 & 63.5 & 16.1 & 1.2 & 4.6 & 35.7 \\ 
CBST~\cite{zou2018unsupervised} & 69.6 & 28.7 & 69.5 & 11.9 & \bf 13.6 & 82.0 & \bf 81.9 & 49.1 & \bf 14.5 & 66.0 & 6.6 & \bf 3.7 & \bf 32.4 & 36.1 \\
AdaSegNet~\cite{tsai2018learning}  & 78.9 & 29.2 & 75.5 & 0.1 & 4.8 & 72.6 & 76.7 & 43.4 & 8.8 & 71.1 & 16.0 & 3.6 & 8.4 & 37.6 \\
SWD (ours)  & \bf 83.3 & \bf 35.4 & \bf 82.1 & \bf 12.2 & 12.6 & \bf 83.8 & 76.5 & 47.4 & 12.0 & \bf 71.5 & \bf 17.9 & 1.6 & 29.7 & \bf 43.5 \\
\midrule
\midrule
Source only (ResNet101) & 55.6 & 23.8 & 74.6 & 6.1 & 12.1 & 74.8 & 79.0 & \bf 55.3 & 19.1 & 39.6 & 23.3 & 13.7 & 25.0 & 38.6\\
AdaSegNet~\cite{tsai2018learning} & 79.2 & \bf 37.2 & 78.8 & 9.9 & 10.5 & 78.2 & \bf 80.5 & 53.5 & \bf 19.6 & 67.0 & 29.5 & \bf 21.6 & 31.3 & 45.9  \\
SWD (ours) & \bf 82.4 & 33.2 & \bf 82.5 & \bf 22.6 & \bf 19.7 & \bf 83.7 & 78.8 & 44.0 & 17.9 & \bf 75.4 & \bf 30.2 & 14.4 & \bf 39.9 & \bf 48.1\\
\bottomrule
  \end{tabular}
  }
  \end{center}
   \vspace{-6mm}
     \caption{Adaptation results from Synthia to Cityscapes. We compare our results with other state-of-the-art approaches that are based on the standard VGG-16 or ResNet-101 backbone.}
      \label{tb:synthia2city}
      \vspace{-3mm}
\end{table*}

%% file: Tables/detection.tex
\begin{table*}[tbp]
\setlength{\tabcolsep}{4pt}
\begin{center}
\scalebox{0.75}{
\begin{tabular}{l|cccccccccccc|c }
\toprule
Method    & plane 	& bcycl 	& bus  	& car  	& horse 	& knife 	& mcycl 	& person 	& plant 	& sktbrd 	& train 	& truck & mAP \\
\midrule
\midrule
Source only & 0.5 & 0.4 & 9.4 & \bf 9.4 & 7.2 & \bf 0.1 & 1.3 & 4.6 & 0.5 & 0.3 & 1.5 & 0.9 & 3.0 \\
MCD~\cite{saito2017maximum} & 11.8 & 1.3 & 10.3 & 3.3 & 10.2 & \bf 0.1 & \bf 8.0 & 2.4 & 0.6 & \bf 1.5 & 5.8 & \bf 1.1 & 4.7  \\
SWD (ours)  & \bf 11.9 & \bf 2.0 & \bf 15.5 & 5.4 & \bf 13.1 & \bf 0.1 & 4.7 & \bf 9.8 & \bf 0.9 & 1.0 & \bf 6.2 & 0.7 & \bf 5.9 \\
\bottomrule
  \end{tabular}
  }
  \end{center}
   \vspace{-6mm}
     \caption{Results of unsupervised domain adaptation on VisDA 2018~\cite{peng2018syn2real} object detection track. This task evaluates the adaptation capability from synthetic CAD model images to real-world MSCOCO images (COCO17-val). We report mean average precision (mAP) at 0.5 IoU using SSD with Inception-V2 backbone. Our method outperforms the direct comparable method MCD~\cite{saito2017maximum} by 25\% relatively.}
      \label{tb:detection}
      \vspace{-3mm}
\end{table*}

%% file: swd.bbl
\begin{thebibliography}{10}\itemsep=-1pt

\bibitem{arjovsky2017wasserstein}
Martin Arjovsky, Soumith Chintala, and L{\'e}on Bottou.
\newblock Wasserstein generative adversarial networks.
\newblock In {\em ICML}, 2017.

\bibitem{ben2010theory}
Shai Ben-David, John Blitzer, Koby Crammer, Alex Kulesza, Fernando Pereira, and
  Jennifer~Wortman Vaughan.
\newblock A theory of learning from different domains.
\newblock {\em Machine learning}, 2010.

\bibitem{ben2007analysis}
Shai Ben-David, John Blitzer, Koby Crammer, and Fernando Pereira.
\newblock Analysis of representations for domain adaptation.
\newblock In {\em NIPS}, 2007.

\bibitem{bonneel2015sliced}
Nicolas Bonneel, Julien Rabin, Gabriel Peyr{\'e}, and Hanspeter Pfister.
\newblock Sliced and radon wasserstein barycenters of measures.
\newblock {\em JMIV}, 2015.

\bibitem{bousmalis2017unsupervised}
Konstantinos Bousmalis, Nathan Silberman, David Dohan, Dumitru Erhan, and Dilip
  Krishnan.
\newblock Unsupervised pixel-level domain adaptation with generative
  adversarial networks.
\newblock In {\em CVPR}, 2017.

\bibitem{bousmalis2016domain}
Konstantinos Bousmalis, George Trigeorgis, Nathan Silberman, Dilip Krishnan,
  and Dumitru Erhan.
\newblock Domain separation networks.
\newblock In {\em NIPS}, 2016.

\bibitem{chen2017no}
Yi-Hsin Chen, Wei-Yu Chen, Yu-Ting Chen, Bo-Cheng Tsai, Yu-Chiang~Frank Wang,
  and Min Sun.
\newblock No more discrimination: Cross city adaptation of road scene
  segmenters.
\newblock In {\em ICCV}, 2017.

\bibitem{cordts2016cityscapes}
Marius Cordts, Mohamed Omran, Sebastian Ramos, Timo Rehfeld, Markus Enzweiler,
  Rodrigo Benenson, Uwe Franke, Stefan Roth, and Bernt Schiele.
\newblock The cityscapes dataset for semantic urban scene understanding.
\newblock In {\em CVPR}, 2016.

\bibitem{courty2017joint}
Nicolas Courty, R{\'e}mi Flamary, Amaury Habrard, and Alain Rakotomamonjy.
\newblock Joint distribution optimal transportation for domain adaptation.
\newblock In {\em NIPS}, 2017.

\bibitem{courty2015optimal}
Nicolas Courty, R{\'e}mi Flamary, Devis Tuia, and Alain Rakotomamonjy.
\newblock Optimal transport for domain adaptation.
\newblock {\em TPAMI}, 2016.

\bibitem{cuturi2013sinkhorn}
Marco Cuturi.
\newblock Sinkhorn distances: Lightspeed computation of optimal transport.
\newblock In {\em NIPS}, 2013.

\bibitem{damodaran2018deepjdot}
Bharath~Bhushan Damodaran, Benjamin Kellenberger, R{\'e}mi Flamary, Devis Tuia,
  and Nicolas Courty.
\newblock Deepjdot: Deep joint distribution optimal transport for unsupervised
  domain adaptation.
\newblock In {\em ECCV}, 2018.

\bibitem{deng2009imagenet}
Jia Deng, Wei Dong, Richard Socher, Li-Jia Li, Kai Li, and Li Fei-Fei.
\newblock Imagenet: A large-scale hierarchical image database.
\newblock In {\em CVPR}, 2009.

\bibitem{deshpande2018generative}
Ishan Deshpande, Ziyu Zhang, and Alexander Schwing.
\newblock Generative modeling using the sliced wasserstein distance.
\newblock In {\em CVPR}, 2018.

\bibitem{french2017self}
Geoffrey French, Michal Mackiewicz, and Mark Fisher.
\newblock Self-ensembling for visual domain adaptation.
\newblock {\em arXiv:1706.05208}, 2017.

\bibitem{frogner2015learning}
Charlie Frogner, Chiyuan Zhang, Hossein Mobahi, Mauricio Araya, and Tomaso~A
  Poggio.
\newblock Learning with a wasserstein loss.
\newblock In {\em NIPS}, 2015.

\bibitem{ganin2014unsupervised}
Yaroslav Ganin and Victor Lempitsky.
\newblock Unsupervised domain adaptation by backpropagation.
\newblock In {\em ICML}, 2014.

\bibitem{ganin2016domain}
Yaroslav Ganin, Evgeniya Ustinova, Hana Ajakan, Pascal Germain, Hugo
  Larochelle, Fran{\c{c}}ois Laviolette, Mario Marchand, and Victor Lempitsky.
\newblock Domain-adversarial training of neural networks.
\newblock {\em JMLR}, 2016.

\bibitem{goodfellow2014generative}
Ian Goodfellow, Jean Pouget-Abadie, Mehdi Mirza, Bing Xu, David Warde-Farley,
  Sherjil Ozair, Aaron Courville, and Yoshua Bengio.
\newblock Generative adversarial nets.
\newblock In {\em NIPS}, 2014.

\bibitem{haeusser2017associative}
Philip Haeusser, Thomas Frerix, Alexander Mordvintsev, and Daniel Cremers.
\newblock Associative domain adaptation.
\newblock In {\em ICCV}, 2017.

\bibitem{haker2004optimal}
Steven Haker, Lei Zhu, Allen Tannenbaum, and Sigurd Angenent.
\newblock Optimal mass transport for registration and warping.
\newblock {\em IJCV}, 2004.

\bibitem{he2016deep}
Kaiming He, Xiangyu Zhang, Shaoqing Ren, and Jian Sun.
\newblock Deep residual learning for image recognition.
\newblock In {\em CVPR}, 2016.

\bibitem{hoffman2017cycada}
Judy Hoffman, Eric Tzeng, Taesung Park, Jun-Yan Zhu, Phillip Isola, Kate
  Saenko, Alexei~A Efros, and Trevor Darrell.
\newblock Cycada: Cycle-consistent adversarial domain adaptation.
\newblock In {\em ICML}, 2018.

\bibitem{hoffman2016fcns}
Judy Hoffman, Dequan Wang, Fisher Yu, and Trevor Darrell.
\newblock Fcns in the wild: Pixel-level adversarial and constraint-based
  adaptation.
\newblock {\em arXiv:1612.02649}, 2016.

\bibitem{huang2018domain}
Haoshuo Huang, Qixing Huang, and Philipp Krahenbuhl.
\newblock Domain transfer through deep activation matching.
\newblock In {\em ECCV}, 2018.

\bibitem{hull1994database}
Jonathan~J. Hull.
\newblock A database for handwritten text recognition research.
\newblock {\em TPAMI}, 1994.

\bibitem{kantorovitch1958translocation}
Leonid Kantorovitch.
\newblock On the translocation of masses.
\newblock {\em Management Science}, 1958.

\bibitem{kingma2014adam}
Diederik~P Kingma and Jimmy Ba.
\newblock Adam: A method for stochastic optimization.
\newblock {\em arXiv:1412.6980}, 2014.

\bibitem{kolouri2017sliced}
Soheil Kolouri, Gustavo~K Rohde, and Heiko Hoffmann.
\newblock Sliced wasserstein distance for learning gaussian mixture models.
\newblock In {\em CVPR}, 2018.

\bibitem{krizhevsky2012imagenet}
Alex Krizhevsky, Ilya Sutskever, and Geoffrey~E Hinton.
\newblock Imagenet classification with deep convolutional neural networks.
\newblock In {\em NIPS}, 2012.

\bibitem{lecun1998gradient}
Yann LeCun, L{\'e}on Bottou, Yoshua Bengio, and Patrick Haffner.
\newblock Gradient-based learning applied to document recognition.
\newblock {\em Proceedings of the IEEE}, 1998.

\bibitem{lee2018wasserstein}
Kwonjoon Lee, Weijian Xu, Fan Fan, and Zhuowen Tu.
\newblock Wasserstein introspective neural networks.
\newblock In {\em CVPR}, 2018.

\bibitem{li2018adaptive}
Yanghao Li, Naiyan Wang, Jianping Shi, Xiaodi Hou, and Jiaying Liu.
\newblock Adaptive batch normalization for practical domain adaptation.
\newblock {\em Pattern Recognition}, 2018.

\bibitem{lin2014microsoft}
Tsung-Yi Lin, Michael Maire, Serge Belongie, James Hays, Pietro Perona, Deva
  Ramanan, Piotr Doll{\'a}r, and C~Lawrence Zitnick.
\newblock Microsoft coco: Common objects in context.
\newblock In {\em ECCV}, 2014.

\bibitem{liu2017unsupervised}
Ming-Yu Liu, Thomas Breuel, and Jan Kautz.
\newblock Unsupervised image-to-image translation networks.
\newblock In {\em NIPS}, 2017.

\bibitem{liu2016coupled}
Ming-Yu Liu and Oncel Tuzel.
\newblock Coupled generative adversarial networks.
\newblock In {\em NIPS}, 2016.

\bibitem{liu2016ssd}
Wei Liu, Dragomir Anguelov, Dumitru Erhan, Christian Szegedy, Scott Reed,
  Cheng-Yang Fu, and Alexander~C Berg.
\newblock Ssd: Single shot multibox detector.
\newblock In {\em ECCV}, 2016.

\bibitem{long2015learning}
Mingsheng Long, Yue Cao, Jianmin Wang, and Michael~I Jordan.
\newblock Learning transferable features with deep adaptation networks.
\newblock In {\em ICML}, 2015.

\bibitem{long2016unsupervised}
Mingsheng Long, Han Zhu, Jianmin Wang, and Michael~I Jordan.
\newblock Unsupervised domain adaptation with residual transfer networks.
\newblock In {\em NIPS}, 2016.

\bibitem{maaten2008visualizing}
Laurens van~der Maaten and Geoffrey Hinton.
\newblock Visualizing data using t-sne.
\newblock {\em JMLR}, 2008.

\bibitem{mi2018variational}
Liang Mi, Wen Zhang, Xianfeng Gu, and Yalin Wang.
\newblock Variational wasserstein clustering.
\newblock In {\em ECCV}, 2018.

\bibitem{moiseev2013evaluation}
Boris Moiseev, Artem Konev, Alexander Chigorin, and Anton Konushin.
\newblock Evaluation of traffic sign recognition methods trained on
  synthetically generated data.
\newblock In {\em ACIVS}. Springer, 2013.

\bibitem{monge1781memoire}
Gaspard Monge.
\newblock M{\'e}moire sur la th{\'e}orie des d{\'e}blais et des remblais.
\newblock {\em Histoire de l'Acad{\'e}mie Royale des Sciences de Paris}, 1781.

\bibitem{moreno2012unifying}
Jose~G Moreno-Torres, Troy Raeder, Roc{\'\i}O Alaiz-Rodr{\'\i}Guez, Nitesh~V
  Chawla, and Francisco Herrera.
\newblock A unifying view on dataset shift in classification.
\newblock {\em Pattern Recognition}, 2012.

\bibitem{murez2018image}
Zak Murez, Soheil Kolouri, David Kriegman, Ravi Ramamoorthi, and Kyungnam Kim.
\newblock Image to image translation for domain adaptation.
\newblock In {\em CVPR}, 2018.

\bibitem{netzer2011reading}
Yuval Netzer, Tao Wang, Adam Coates, Alessandro Bissacco, Bo Wu, and Andrew~Y
  Ng.
\newblock Reading digits in natural images with unsupervised feature learning.
\newblock In {\em NIPS workshop}, 2011.

\bibitem{pan2010survey}
Sinno~Jialin Pan, Qiang Yang, et~al.
\newblock A survey on transfer learning.
\newblock {\em IEEE Transactions on knowledge and data engineering}, 2010.

\bibitem{pedregosa2011scikit}
Fabian Pedregosa, Ga{\"e}l Varoquaux, Alexandre Gramfort, Vincent Michel,
  Bertrand Thirion, Olivier Grisel, Mathieu Blondel, Peter Prettenhofer, Ron
  Weiss, Vincent Dubourg, et~al.
\newblock Scikit-learn: Machine learning in python.
\newblock {\em JMLR}, 2011.

\bibitem{peng2018zero}
Kuan-Chuan Peng, Ziyan Wu, and Jan Ernst.
\newblock Zero-shot deep domain adaptation.
\newblock In {\em ECCV}, 2018.

\bibitem{visda2017}
Xingchao Peng, Ben Usman, Neela Kaushik, Judy Hoffman, Dequan Wang, and Kate
  Saenko.
\newblock Visda: The visual domain adaptation challenge.
\newblock {\em arXiv:1710.06924}, 2017.

\bibitem{peng2018syn2real}
Xingchao Peng, Ben Usman, Kuniaki Saito, Neela Kaushik, Judy Hoffman, and Kate
  Saenko.
\newblock Syn2real: A new benchmark for synthetic-to-real visual domain
  adaptation.
\newblock {\em arXiv:1806.09755}, 2018.

\bibitem{pitie2005n}
Francois Pitie, Anil~C Kokaram, and Rozenn Dahyot.
\newblock N-dimensional probablility density function transfer and its
  application to colour transfer.
\newblock In {\em ICCV}, 2005.

\bibitem{rabin2011wasserstein}
Julien Rabin, Gabriel Peyr{\'e}, Julie Delon, and Marc Bernot.
\newblock Wasserstein barycenter and its application to texture mixing.
\newblock In {\em SSVM}, 2011.

\bibitem{richter2016playing}
Stephan~R Richter, Vibhav Vineet, Stefan Roth, and Vladlen Koltun.
\newblock Playing for data: Ground truth from computer games.
\newblock In {\em ECCV}, 2016.

\bibitem{ros2016synthia}
German Ros, Laura Sellart, Joanna Materzynska, David Vazquez, and Antonio~M
  Lopez.
\newblock The synthia dataset: A large collection of synthetic images for
  semantic segmentation of urban scenes.
\newblock In {\em CVPR}, 2016.

\bibitem{rubner1998metric}
Yossi Rubner, Carlo Tomasi, and Leonidas~J Guibas.
\newblock A metric for distributions with applications to image databases.
\newblock In {\em ICCV}, 1998.

\bibitem{saenko2010adapting}
Kate Saenko, Brian Kulis, Mario Fritz, and Trevor Darrell.
\newblock Adapting visual category models to new domains.
\newblock In {\em ECCV}, 2010.

\bibitem{saito2017maximum}
Kuniaki Saito, Kohei Watanabe, Yoshitaka Ushiku, and Tatsuya Harada.
\newblock Maximum classifier discrepancy for unsupervised domain adaptation.
\newblock In {\em CVPR}, 2018.

\bibitem{saito2018open}
Kuniaki Saito, Shohei Yamamoto, Yoshitaka Ushiku, and Tatsuya Harada.
\newblock Open set domain adaptation by backpropagation.
\newblock In {\em ECCV}, 2018.

\bibitem{sankaranarayanan2017generate}
Swami Sankaranarayanan, Yogesh Balaji, Carlos~D Castillo, and Rama Chellappa.
\newblock Generate to adapt: Aligning domains using generative adversarial
  networks.
\newblock In {\em CVPR}, 2018.

\bibitem{sankaranarayanan2018learning}
Swami Sankaranarayanan, Yogesh Balaji, Arpit Jain, Ser Nam~Lim, and Rama
  Chellappa.
\newblock Learning from synthetic data: Addressing domain shift for semantic
  segmentation.
\newblock In {\em CVPR}, 2018.

\bibitem{shimodaira2000improving}
Hidetoshi Shimodaira.
\newblock Improving predictive inference under covariate shift by weighting the
  log-likelihood function.
\newblock {\em Journal of statistical planning and inference}, 2000.

\bibitem{shrivastava2017learning}
Ashish Shrivastava, Tomas Pfister, Oncel Tuzel, Joshua Susskind, Wenda Wang,
  and Russell Webb.
\newblock Learning from simulated and unsupervised images through adversarial
  training.
\newblock In {\em CVPR}, 2017.

\bibitem{shu2018dirt}
Rui Shu, Hung~H Bui, Hirokazu Narui, and Stefano Ermon.
\newblock A dirt-t approach to unsupervised domain adaptation.
\newblock In {\em ICLR}, 2018.

\bibitem{simonyan2014very}
Karen Simonyan and Andrew Zisserman.
\newblock Very deep convolutional networks for large-scale image recognition.
\newblock In {\em ICLR}, 2015.

\bibitem{stallkamp2011german}
Johannes Stallkamp, Marc Schlipsing, Jan Salmen, and Christian Igel.
\newblock The german traffic sign recognition benchmark: a multi-class
  classification competition.
\newblock In {\em IJCNN}, 2011.

\bibitem{storkey2009training}
Amos Storkey.
\newblock When training and test sets are different: characterizing learning
  transfer.
\newblock {\em Dataset shift in machine learning}, 2009.

\bibitem{sundermeyer2018implicit}
Martin Sundermeyer, Zoltan-Csaba Marton, Maximilian Durner, Manuel Brucker, and
  Rudolph Triebel.
\newblock Implicit 3d orientation learning for 6d object detection from rgb
  images.
\newblock In {\em ECCV}, 2018.

\bibitem{szegedy2016rethinking}
Christian Szegedy, Vincent Vanhoucke, Sergey Ioffe, Jon Shlens, and Zbigniew
  Wojna.
\newblock Rethinking the inception architecture for computer vision.
\newblock In {\em CVPR}, 2016.

\bibitem{tsai2018learning}
Y.-H. Tsai, W.-C. Hung, S. Schulter, K. Sohn, M.-H. Yang, and M. Chandraker.
\newblock Learning to adapt structured output space for semantic segmentation.
\newblock In {\em CVPR}, 2018.

\bibitem{tzeng2017adversarial}
Eric Tzeng, Judy Hoffman, Kate Saenko, and Trevor Darrell.
\newblock Adversarial discriminative domain adaptation.
\newblock In {\em CVPR}, 2017.

\bibitem{tzeng2014deep}
Eric Tzeng, Judy Hoffman, Ning Zhang, Kate Saenko, and Trevor Darrell.
\newblock Deep domain confusion: Maximizing for domain invariance.
\newblock {\em arXiv:1412.3474}, 2014.

\bibitem{villani2009optimal}
Cedric Villani.
\newblock Optimal transport, old and new.
\newblock {\em Springer-Verlag}, 2009.

\bibitem{wu2018wasserstein}
Jiqing Wu, Zhiwu Huang, Janine Thoma, Dinesh Acharya, and Luc Van~Gool.
\newblock Wasserstein divergence for gans.
\newblock In {\em ECCV}, 2018.

\bibitem{wu2018dcan}
Zuxuan Wu, Xintong Han, Yen-Liang Lin, Mustafa~Gkhan Uzunbas, Tom Goldstein,
  Ser~Nam Lim, and Larry~S Davis.
\newblock Dcan: Dual channel-wise alignment networks for unsupervised scene
  adaptation.
\newblock In {\em ECCV}, 2018.

\bibitem{zellinger2017central}
Werner Zellinger, Thomas Grubinger, Edwin Lughofer, Thomas Natschl{\"a}ger, and
  Susanne Saminger-Platz.
\newblock Central moment discrepancy (cmd) for domain-invariant representation
  learning.
\newblock In {\em ICLR}, 2017.

\bibitem{zhang2017curriculum}
Yang Zhang, Philip David, and Boqing Gong.
\newblock Curriculum domain adaptation for semantic segmentation of urban
  scenes.
\newblock In {\em ICCV}, 2017.

\bibitem{zhao2017pyramid}
Hengshuang Zhao, Jianping Shi, Xiaojuan Qi, Xiaogang Wang, and Jiaya Jia.
\newblock Pyramid scene parsing network.
\newblock In {\em CVPR}, 2017.

\bibitem{zou2018unsupervised}
Yang Zou, Zhiding Yu, BVK~Vijaya Kumar, and Jinsong Wang.
\newblock Unsupervised domain adaptation for semantic segmentation via
  class-balanced self-training.
\newblock In {\em ECCV}, 2018.

\end{thebibliography}
